\documentclass[10pt,twocolumn,letterpaper]{article}

\usepackage{iccv}
\usepackage{times}
\usepackage{epsfig}
\usepackage{graphicx}
\usepackage{amsmath}
\usepackage{amssymb}

\usepackage{booktabs}
\usepackage{multirow}
\usepackage{mathtools}
\usepackage{bm}
\usepackage{array}
\makeatletter
\@namedef{ver@everyshi.sty}{}
\makeatother
\usepackage{tikz}

\DeclareMathOperator*{\argmax}{arg\,max}

\newcommand{\mat}[1]{\mathbf{#1}}
\newcommand{\matsym}[1]{\bm{#1}}
\renewcommand{\vec}[1]{\mathbf{#1}}
\newcommand{\vecsym}[1]{\bm{#1}}

\newcommand{\se}{\mathfrak{se}}

\newcommand{\SEthree}{\ensuremath{\mathrm{SE}(3)}\xspace}

\newcolumntype{P}[1]{>{\centering\arraybackslash}p{#1}}

\usepackage[breaklinks=true,bookmarks=false]{hyperref}

\iccvfinalcopy % *** Uncomment this line for the final submission

\def\iccvPaperID{5672} % *** Enter the ICCV Paper ID here

\ificcvfinal\fi

\newcommand\copyrighttext{%
	\footnotesize \textcopyright 2020 IEEE. Personal use of this material is permitted.
	Permission from IEEE must be obtained for all other uses, in any current or future
	media, including reprinting/republishing this material for advertising or promotional
	purposes, creating new collective works, for resale or redistribution to servers or
	lists, or reuse of any copyrighted component of this work in other works.
	Presented at the 2019 IEEE/CVF International Conference on Computer Vision (ICCV).
	DOI: 10.1109/ICCV.2019.00596.
	Publisher version: https://ieeexplore.ieee.org/document/9010932.}
\newcommand\copyrightnotice{%
	\begin{tikzpicture}[remember picture,overlay]
	\node[anchor=south west,yshift=10pt,xshift=1.6cm] at (current page.south west) {\parbox{\textwidth}{\copyrighttext}};
	\end{tikzpicture}%
}

\begin{document}

%%%%%%%%% TITLE
\title{EM-Fusion: Dynamic Object-Level SLAM With Probabilistic Data Association}

\author{Michael Strecke and J\"org St\"uckler\\
Embodied Vision Group, Max Planck Institute for Intelligent Systems\\
{\tt\small \{michael.strecke,joerg.stueckler\}@tue.mpg.de}
}

\maketitle
\copyrightnotice
\ificcvfinal\thispagestyle{empty}\fi

%%%%%%%%% ABSTRACT
\begin{abstract}
   The majority of approaches for acquiring dense 3D environment maps with RGB-D cameras assumes static environments or rejects moving objects as outliers. 
The representation and tracking of moving objects, however, has significant potential for applications in robotics or augmented reality.
In this paper, we propose a novel approach to dynamic SLAM with dense object-level representations.
We represent rigid objects in local volumetric signed distance function (SDF) maps, and formulate multi-object tracking as direct alignment of RGB-D images with the SDF representations.
Our main novelty is a probabilistic formulation which naturally leads to strategies for data association and occlusion handling.
We analyze our approach in experiments and demonstrate that our approach compares favorably with the state-of-the-art methods in terms of robustness and accuracy.

\end{abstract}

%%%%%%%%% BODY TEXT
\section{Introduction}

\begin{figure*}[h!]
	\begin{center}
	\includegraphics[width=\linewidth]{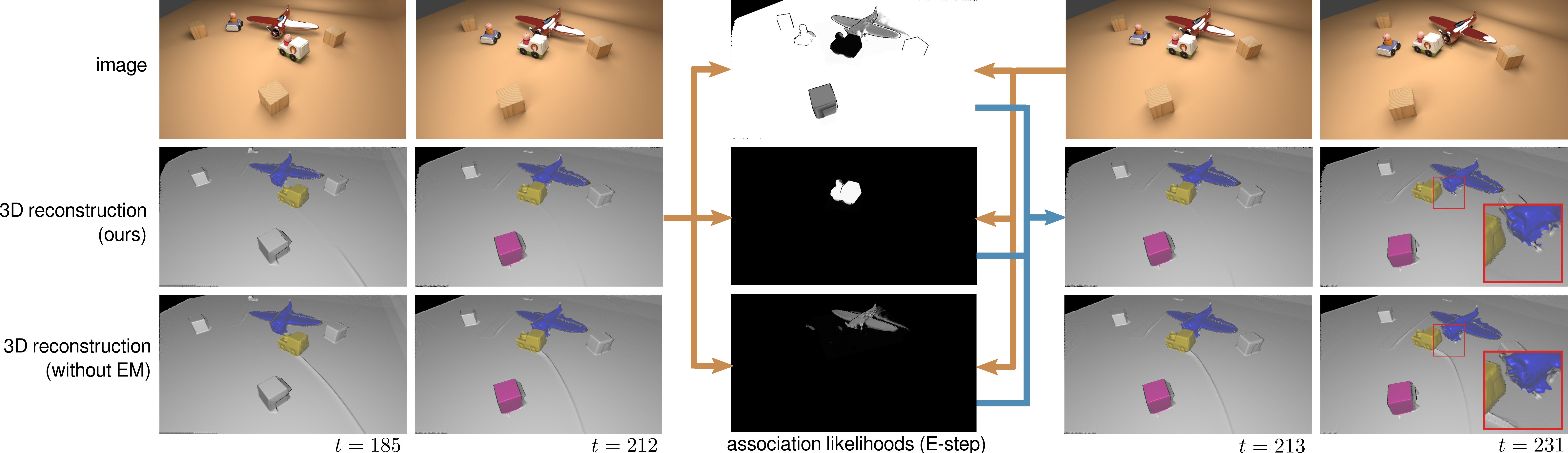}
	\end{center}
	\caption{Dynamic object-level SLAM with probabilistic data association. We infer the association likelihood of pixels with objects in an expectation-maximization framework. The probabilistic data association improves accuracy and robustness of tracking and mapping. It implicitly handles occlusions. The E-step estimates the association likelihoods based on the data likelihood of the current image given the latest object maps and poses. In the M-step poses and map are updated with the measurements according to the association likelihoods. Association likelihoods are visualized for the background (top), the train (middle) and the airplane (bottom). The moving train occludes the table and the airplane which is well recovered by the association likelihoods. Without association likelihoods, artifacts are integrated into the map due to wrong data association.}
	\label{fig:teaser}
\end{figure*}

RGB-D cameras are popular devices for dense visual 3D scene acquisition. 
Most approaches to simultaneous localization and mapping (SLAM) with RGB-D cameras only map the static part of the environment and localize the camera within this map.
While some approaches filter dynamic objects as outliers from the measurements, SLAM of multiple moving objects has attracted only little attention so far. 
In many applications of robotics and augmented reality (AR), however, agents interact with the environment and hence the environment state is dynamic.
Approaches that concurrently track multiple moving objects hence have rich potential for robotics and AR applications. 

In this paper, we propose a novel approach to dynamic SLAM that maps and tracks objects in the scene. 
We detect objects through instance segmentation of the images and subsequently perform tracking and mapping of the static background and the objects.
In previous approaches~\cite{ruenz2017_cofusion,ruenz2018_maskfusion,xu2019_midfusion}, data association of measurements to objects is either solved through image-based instance segmentation or by raycasting in the maps. 
We propose to determine the unknown association of pixels to objects in a probabilistic expectation maximization (EM~\cite{bishop2006_prm}) formulation which estimates the soft association likelihood from the likelihood of the measurements in our map representation.
The probabilistic association provides additional geometric cues and implicitly handles occlusions for object segmentation, tracking, and mapping (see Fig.~\ref{fig:teaser}).
We represent the object maps by volumetric signed distance functions (SDFs).
We augment the maximum likelihood integration of the SDF from depths to incorporate their association likelihood.
The probabilistic data association facilitates the direct alignment of the depth maps with the SDF object maps.
This avoids projective data association through raycasting which is needed for the ICP algorithm. 
In our experiments, we evaluate our approach on several datasets and demonstrate superior performance over the state-of-the-art methods.
Our results demonstrate that proper probabilistic treatment of data associations is a key ingredient to robust object-level SLAM in dynamic scenes.
In summary, we make the following contributions in our work,
\begin{itemize}
\setlength\itemsep{0.1em}
	\item We propose a probabilistic EM formulation for dynamic object-level SLAM that naturally leads to data association and occlusion handling strategies.
	\item Based on our EM formulation, we approach multi-object tracking as direct alignment of RGB-D images with SDF object representations and evaluate this tracking approach for dense dynamic SLAM.
	\item Our approach achieves state-of-the-art performance on several datasets for dynamic object-level SLAM. 
\end{itemize}

\section{Related work}
{\bf Static SLAM:} Simultaneous localization and mapping (SLAM) with RGB-D sensors has seen tremendous progress quickly after the sensors have become broadly available on the market.
KinectFusion~\cite{newcombe2011_kinectfusion} is a prominent approach that incrementally tracks the camera motion and maps the environment densely in volumetric signed distance function (SDF) grids.
Several other RGB-D SLAM approaches have been proposed that differ in tracking methods such as ICP~\cite{newcombe2011_kinectfusion}, direct image alignment~\cite{kerl2013_dvoslam} or SDF alignment~\cite{bylow2013_sdftracking}, and map representations such as surfels~\cite{keller13_pointbasedfusion} or keyframes~\cite{kerl2013_dvoslam}.
Extensive research has gone into scaling the approaches to large environments~\cite{whelan2012_kintinuous,niessner2013_voxelhashing} or supporting loop-closing~\cite{kerl2013_dvoslam,whelan2015_elasticfusion} to reduce drift.
Some approaches also consider the creation of object-level maps~\cite{salas-moreno2013_slampp,mccormac2018_fusionpp}, but assume the objects to remain static.

{\bf Dynamic SLAM:} Research on tracking and reconstruction of articulated objects such as human body parts~\cite{taylor2016_articulatedhandtracking,tzionas2016_riggedmodels} or robots~\cite{schmidt2015_riggedtracking,cifuentes2017_articulatedtracking} is related to dynamic SLAM.
Recently, some RGB-D SLAM methods have been proposed that represent and track moving rigid objects.
An early approach extends keyframe-based RGB-D SLAM to object-level dynamic SLAM~\cite{stueckler-ijcai13}.
The approach segments moving objects between RGB-D frames~\cite{stueckler-ijcv15} and builds a keyframe pose graph for associated motion segments in the keyframes.
Co-Fusion~\cite{ruenz2017_cofusion} extends surfel-based representations for moving objects.
It combines geometric with motion segmentation to detect moving objects. 
Tracking camera motion with respect to the scene background and the objects is based on ICP alignment using geometry and color cues.
MaskFusion~\cite{ruenz2018_maskfusion} does not use motion segmentation but fuses geometric with a deep-learning based instance segmentation (Mask R-CNN~\cite{he2017_maskrcnn}).
MID-Fusion~\cite{xu2019_midfusion} follows a similar approach, but represents the 3D map in volumetric SDFs using octrees.
We also represent objects in using SDFs but formulate tracking using efficient but accurate direct SDF alignment \cite{bylow2013_sdftracking}.
We also propose novel strategies for handling occlusions and disocclusions.

\section{Proposed Method}
Our dynamic SLAM approach performs incremental tracking and mapping of objects and the static background.
We propose a probabilistic formulation for tracking and mapping of multiple objects which naturally leads to a principled method for data association and occlusion handling.
We represent the 3D shape of objects and background in volumetric SDF representations which we estimate from depth images.
New object instances are initially detected and segmented using a semantic appearance-based deep learning approach (Mask R-CNN~\cite{he2017_maskrcnn}).

\subsection{Probabilistic Dynamic Tracking and Mapping}

We formulate SLAM as maximum likelihood estimation of the camera trajectory and the map from visual observations~$\vec{z}_t$ (the depth images). 
The map is composed of separate TSDF volumes $\vec{m} := \left\{ \vec{m}_i \right\}_{i=0}^N$ for the background ($\vec{m}_0$) and $N$ objects. 
In each camera frame at time~$t$, we track the camera pose with regard to the objects and background with distinct poses~$\vecsym{\xi}_t := \left\{ \vecsym{\xi}_{t,i} \right\}_{i=0}^N$, $\vecsym{\xi}_{t,i} \in \SEthree$. 
We choose incremental tracking and mapping in which we optimize the joint posterior likelihood of the map and the camera poses in the current frame, given all images so far, 
\begin{multline}
	\argmax_{\vec{m}, \vecsym{\xi}_t} p( \vec{m}, \vecsym{\xi}_t \mid \vec{z}_{1:t} ) =\\
	\argmax_{\vec{m}, \vecsym{\xi}_t} p( \vec{z}_t \mid \vec{m}, \vecsym{\xi}_t ) \, p( \vec{m} \mid \vec{z}_{1:t-1} ) \, p( \vecsym{\xi}_t ).
\end{multline}
We optimize the posterior separately first for the camera pose, then for the map.
By causality, each pixel measurement can only be attributed to one of the objects or the background, such that we also need to find the association of each pixel~$u$ to one of the objects.
This association is a latent variable~$c_t = \left\{ c_{t,u} \right\}, c_{t,u} \in \left\{ 0, \ldots, N \right\}$ in our probabilistic model which we infer during the tracking and mapping.

\subsection{Expectation Maximization Framework}

Expectation-maximization (EM) is a natural framework for our problem of finding the latent data association with the map and camera pose estimates.
In EM, we treat the map and camera poses as parameters~$\vecsym{\theta}$ to be optimized.
In the E-step, we recover a variational approximation of the association likelihood given the current parameter estimate from the previous EM iteration,
\begin{equation}
q( c_t ) \leftarrow \argmax_{q(c_t)} \sum_{c_t} q( c_t ) \ln p( \vec{z}_t, c_t \mid \vecsym{\theta} ).
\end{equation}
The maximum is achieved for $q( c_t ) = p( c_t \mid \vec{z}_t, \vecsym{\theta} )$.
For the M-step, we maximize the expected log posterior under the approximate association likelihood
\begin{equation}
\label{eq:mstep}
	\vecsym{\theta} \leftarrow \argmax_{\vecsym{\theta}} \sum_{c_t} q( c_t ) \ln p( \vec{z}_t, c_t \mid \vecsym{\theta} ) + \ln p( \vecsym{\theta} ).
\end{equation}
Note that $p( \vecsym{\theta} ) = p( \vec{m} \mid \vec{z}_{1:t-1} ) \, p( \vecsym{\xi}_t )$.

In our case the E-step can be performed by evaluating
\begin{equation}
	p( c_t \mid \vec{z}_t, \vecsym{\theta} ) = \frac{p( \vec{z}_t \mid c_t, \vecsym{\theta} ) p( c_t \mid \vecsym{\theta} )}{\sum_{c'_t} p( \vec{z}_t \mid c'_t, \theta ) p( c'_t \mid \vecsym{\theta} ) }.
\end{equation}
Since we treat data and association likelihood stochastically independent between pixels, the association likelihood can be determined for each pixel individually.
Assuming uniform prior association likelihood, we arrive at
\begin{equation}
\label{eq:association_posterior}
	p( c_t \mid \vec{z}_t, \vecsym{\theta} ) = \frac{p( \vec{z}_t \mid c_t, \vecsym{\theta} )}{\sum_{c'_t} p( \vec{z}_t \mid c'_t, \vecsym{\theta} )}.
\end{equation}

The M-step is solved individually per object by taking into account the association likelihood of the pixels to the objects.
We optimize first for the camera poses in the previous map and then integrate the measurement into the map using the new pose estimates. 
In the following, we detail the steps in our pipeline that implement the EM algorithm.

\subsection{Image Preprocessing and Projection}

We apply a bilateral filter on the raw depth images to smoothen depth quantization artifacts.
From the filtered depth maps~$D$ we compute 3D point coordinates $\vec{p} = \pi^{-1}( \vec{u}, D(\vec{u}) ) \in \mathbb{R}^3$ at each pixel~$\vec{u} \in \mathbb{R}^2$, where we define $\pi^{-1}( \vec{u}, D(\vec{u}) ) := D(\vec{u}) \, \mat{C}^{-1} \left( u_x, u_y, 1 \right)^\top$ and~$\mat{C}$ is the camera intrinsics matrix of the calibrated pinhole camera.

\subsection{Map Representation}

We represent background and objects maps by volumetric SDFs.
The SDF~$\psi( \vec{p} ): \mathbb{R}^3 \rightarrow \mathbb{R}$ yields the signed distance of a point~$\vec{p}$ to the closest surface represented by the SDF. 
The object surface is determined by the zero level-set $\left\{ \vec{p} \in \mathbb{R}^3: \psi(\vec{p}) = 0 \right\}$ of the SDF.
We implement the volumetric SDF through discretization in a 3D grid of voxels.
The SDF value at a point within the grid is found through trilinear interpolation.
We maintain several SDF volumes: one background volume (resolution $512^3$) and several smaller SDF volumes, one for each detected object (initialized with a size of $64^3$ and resized as needed, s.~Sec,~\ref{sec:instance_detection}).

\subsection{Instance Detection and Segmentation}\label{sec:instance_detection}

For instance detection and segmentation we mostly follow~\cite{mccormac2018_fusionpp}, but adapt the approach for dynamic scenes. 
As in~\cite{mccormac2018_fusionpp} we use Mask R-CNN~\cite{he2017_maskrcnn} to detect and segment object instances.
The Mask R-CNN detector runs at a lower processing rate (sequentially every 30 frames) than the remaining SLAM pipeline, hence, we only have detections available for a subset of frames.
If a detection result is available, we match the detections with the current objects in the map and create new objects for unmatched detections.

Similar to \cite{mccormac2018_fusionpp} we recursively estimate the foreground probability~$p_{\mathit{fg}}( \vec{p} \mid i ) = \mathit{Fg}_i( \vec{p} ) / (\mathit{Fg}_i( \vec{p} ) + \mathit{Bg}_i( \vec{p} ))$ of points~$\vec{p}$ through counts in the corresponding voxels.
The foreground and background counts~$\mathit{Fg}_i( v )$ and~$\mathit{Bg}_i( v )$ of each voxel~$v$ are updated using the associated segments,
\begin{equation}
\label{eq:fg_bg_update}
\begin{split}
\mathit{Fg}_i( v ) &\leftarrow \mathit{Fg}_i( v ) + p^{\mathit{MRCNN}}_{\mathit{fg}}( v )\\
\mathit{Bg}_i( v ) &\leftarrow \mathit{Bg}_i( v ) + \left( 1 - p^{\mathit{MRCNN}}_{\mathit{fg}}( v ) \right)\\
\end{split}
\end{equation} 
The voxels are projected into the image to determine the segmentation likelihood~$p^{\mathit{MRCNN}}_{\mathit{fg}}( v )$ in the associated segment from Mask R-CNN.
During raycasting for visualization and generation of model masks, a point $\vec{p}$ from object $i$ is only rendered if $p_{\mathit{fg}}( \vec{p} \mid i ) > 0.5$ and there is no other model along that ray with a shorter ray distance.
To account for possible occlusions, we only perform the update in \eqref{eq:fg_bg_update} in unoccluded regions, i.e., where the projected mask of the object volume fits the fused segmentation from all objects.

For matching detections with objects, we find the reprojected segmentations of the objects in the map within the current image using raycasting.
We determine the overlap of the reprojected segmentations with the detected segments by the intersection-over-union (IoU) measure.
Segments are associated if their IoU is largest and above a threshold ($0.2$ in our experiments).
Similar to~\cite{mccormac2018_fusionpp}, unmatched segments are used to create new objects by calculating the $10^{\mathit{th}}$ and $90^{\mathit{th}}$ percentiles of the pointcloud generated from the depth image masked by a segment and using them to determine the volume center $\vec{c}_i$ and size $s_i$ (see \cite{mccormac2018_fusionpp} for details).
We choose a padding factor of $2.0$ around these percentiles for the volume size and set the initial volume resolution $r_i$ to $64$ along each axis, yielding a voxel size of $v_i = \frac{s_i}{r_i}$.
If new detections matched with an existing model fall outside the existing volume, it is resized by determining an increased $r_i$ required to fit the new detection and shifting $\vec{c}_i$ by a multiple of $v_i$ so that it is still in the center of the volume.

The new volume is only initialized if its center $\vec{c}_i$ is within 5\,m from the camera and the volumetric IoU with any other volume is lower than $0.5$.
Since Mask R-CNN can deliver false detections, we follow~\cite{mccormac2018_fusionpp} and maintain an existence probability $p_\mathit{ex}(i) = \mathit{Ex}(i) / (\mathit{Ex}(i) + \mathit{NonEx}(i))$, where for each frame with a Mask R-CNN segmentation available $\mathit{Ex}(i)$ is incremented if the object is matched to a segment and otherwise $\mathit{NonEx}(i)$ is incremented.
We delete objects where $p_\mathit{ex}(i) < 0.1$.

\begin{figure}
	\begin{center}
	\setlength{\tabcolsep}{1pt}
	\renewcommand{\arraystretch}{0.6}
	\begin{tabular}{p{0.3\linewidth}p{0.3\linewidth}p{0.3\linewidth}}
	\includegraphics[width=\linewidth]{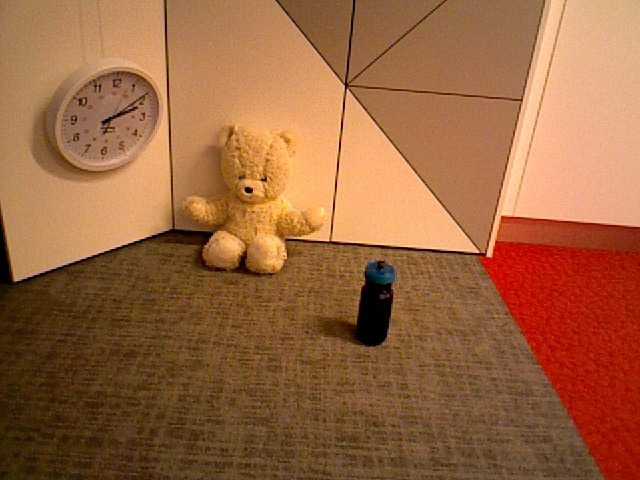} & \includegraphics[width=\linewidth]{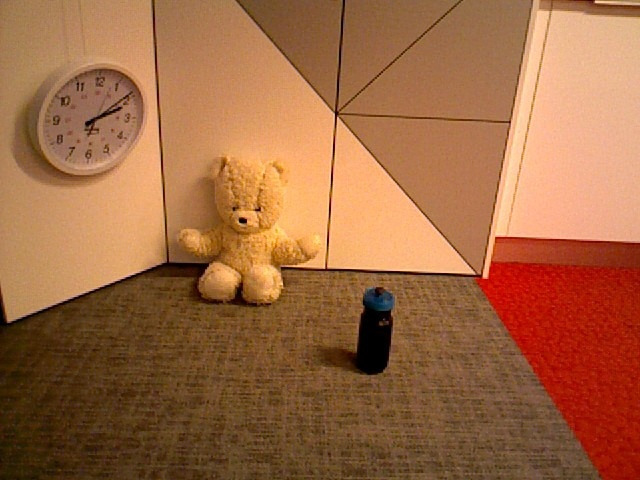} & \includegraphics[width=\linewidth]{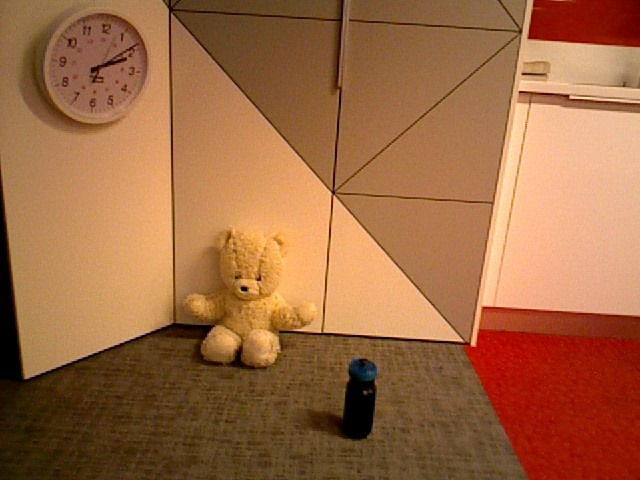} \\
	\includegraphics[width=\linewidth]{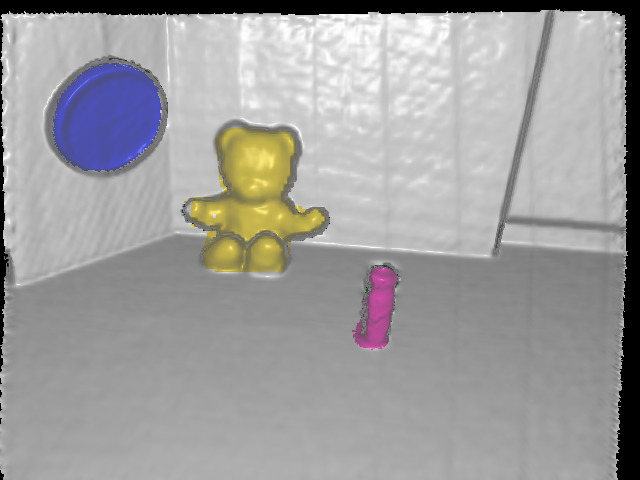} & \includegraphics[width=\linewidth]{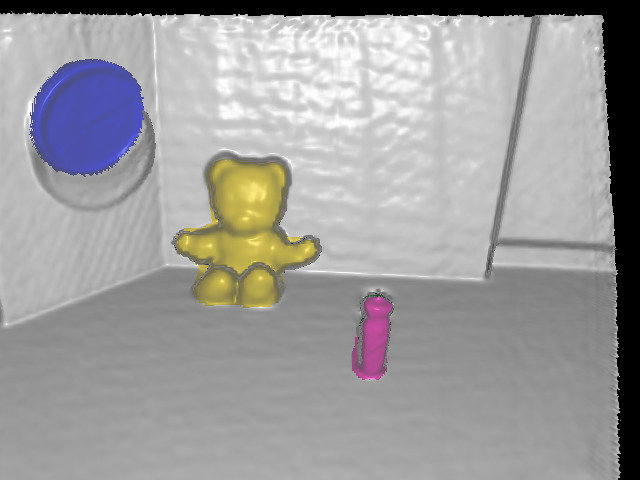} & \includegraphics[width=\linewidth]{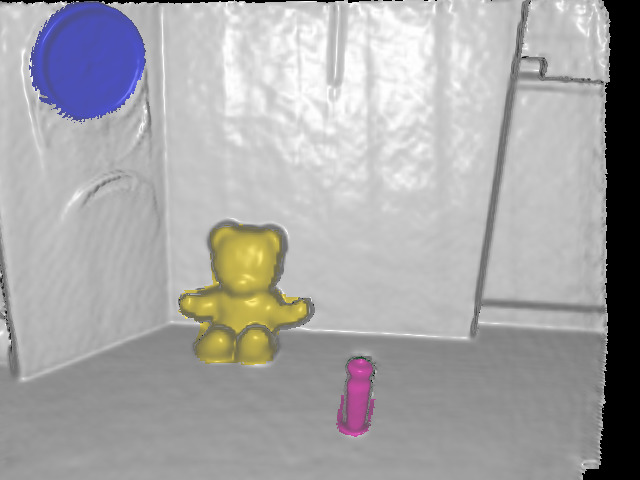} \\
	\includegraphics[width=\linewidth]{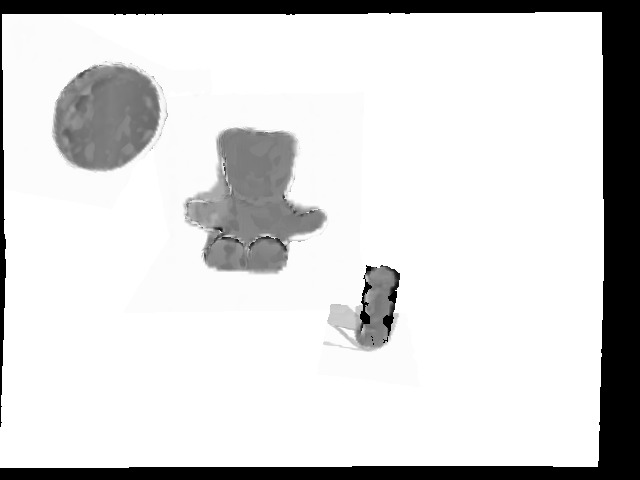} & \includegraphics[width=\linewidth]{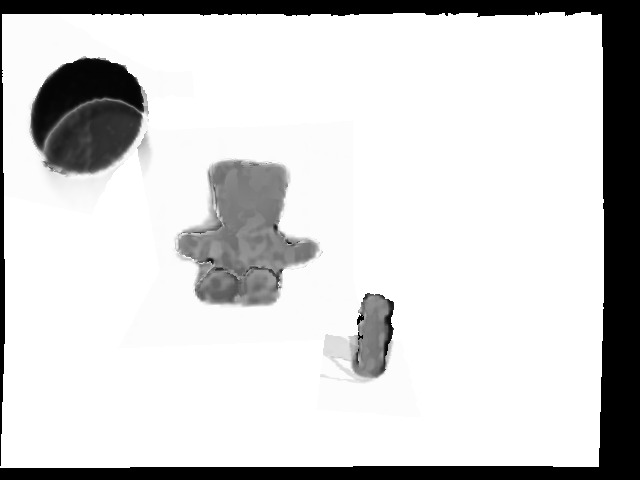} & \includegraphics[width=\linewidth]{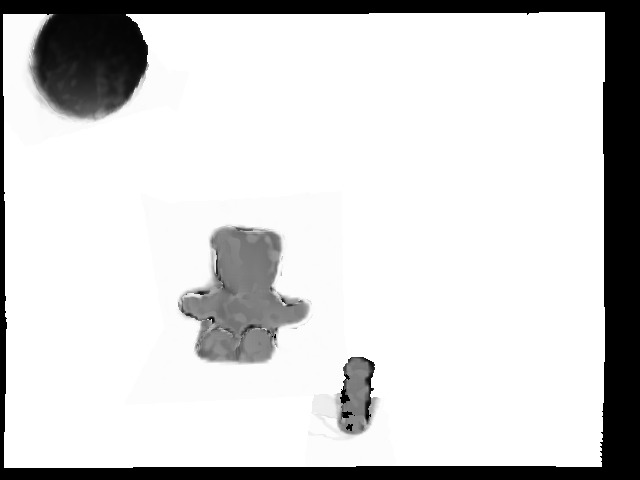} \\
	\includegraphics[width=\linewidth]{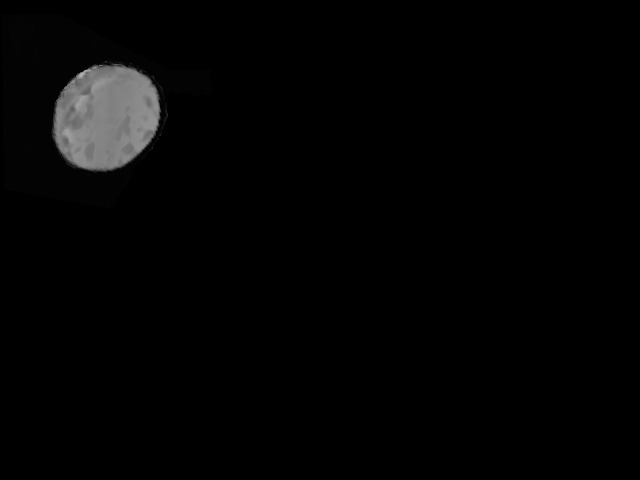} & \includegraphics[width=\linewidth]{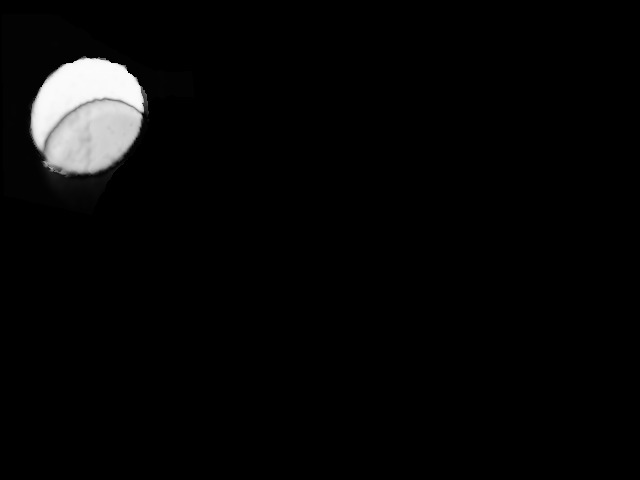} & \includegraphics[width=\linewidth]{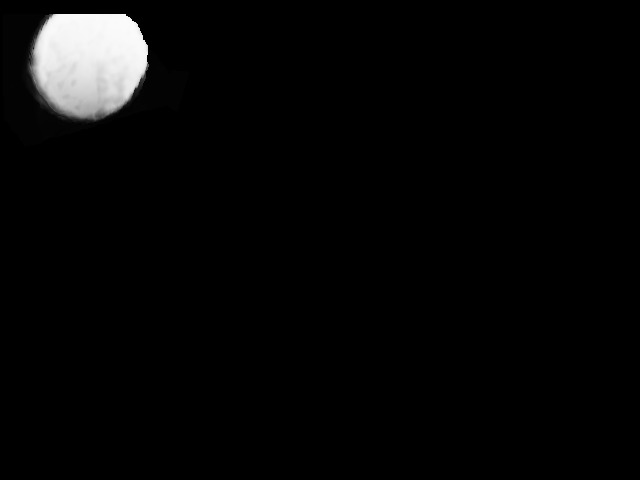} \\
	\end{tabular}
	\end{center}
	\caption{Pixel association likelihood. The E-step of our EM method determines the association likelihood (black: 0, white: 1) for the background (third row) and all objects (fourth row: clock). The association likelihood is determined from the data likelihood of the pixels in all objects given the current pose and map estimates (second row, object segments overlaid by color). Before the clock starts to move, the association weight is equally distributed between the background and the clock model. While the clock moves upwards, the background above the clock becomes occluded and the clock measurements are stronger associated with the object map than with the background.}
	\label{fig:assocweights}
\end{figure}

\subsection{Data Association}

We associate the pixels~$\vec{u}$ in the current frame according to Eq.~\eqref{eq:association_posterior}.
Let~$\vec{p}_i := \mat{T}( \vecsym{\xi}_i ) \, \overline{\pi^{-1}}( \vec{u}, D(\vec{u}) )$ be the local point coordinate of pixel~$\vec{u}$ in the coordinate frame of object~$i$, where we denote~$\overline{\vec{p}} := \left( \vec{p}^\top, 1 \right)^\top$.
We model the data likelihood of a pixel that falls inside the map volume of object $c_t$ with a mixture distribution,
\begin{multline}
	\label{eq:data_association}
	p( \vec{u} \mid c_t, \vecsym{\theta} ) = \alpha \frac{1}{2\sigma} \exp\left( -\frac{|\psi_{c_t}( \vec{p}_{c_t} )|}{\sigma} \right) \, p_{\mathit{fg}}( \vec{p}_{c_t} \mid c_t ) + \\  (1-\alpha) \, p_\mathcal{U}( \vec{p}_{c_t} ),
\end{multline}
where~$\psi_{c_t}$ is the SDF of object~$c_t$.
The mixture is composed of a Laplace distribution which explains the measurement within the object, and a uniform component~$p_\mathcal{U}$ that models outlier measurements and objects that are not yet detected and missing in the multi-object map.
If the pixel is not within the map volume of object~$c_t$, we set its data likelihood to zero for this object.
Hence, the association likelihood is $p( c_t \mid \vec{u}, \vecsym{\theta} ) = \frac{p( \vec{u} \mid c_t, \vecsym{\theta} )}{\sum_{c'_t} p( \vec{u} \mid c'_t, \vecsym{\theta} )}$.

Occlusions are implicitly handled by our data association approach.
If an object is occluded by another object in the map, the association likelihood will be higher within the occluding object.
This results in a lower weight for the measurements in the occluded object for tracking and map integration.
Fig.~\ref{fig:assocweights} illustrates such a case for a clock which is moved upwards along a wall.

\subsection{Tracking}

\begin{figure}
	\begin{center}
	\setlength{\tabcolsep}{1pt}
	\renewcommand{\arraystretch}{0.6}
	\begin{tabular}{p{0.32\linewidth}p{0.32\linewidth}p{0.32\linewidth}}
	\includegraphics[width=\linewidth]{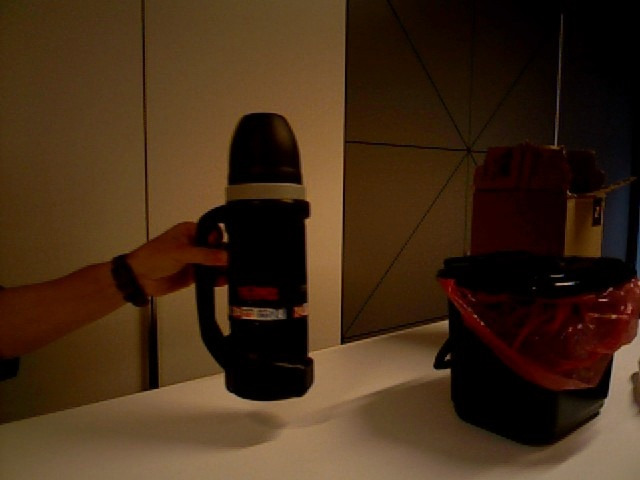} & \includegraphics[width=\linewidth]{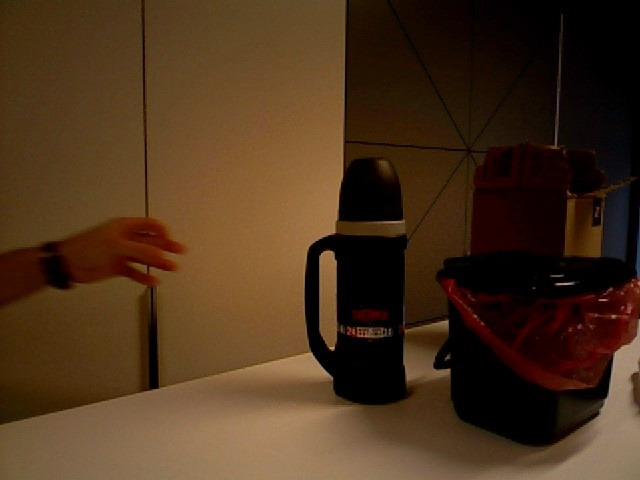} & \includegraphics[width=\linewidth]{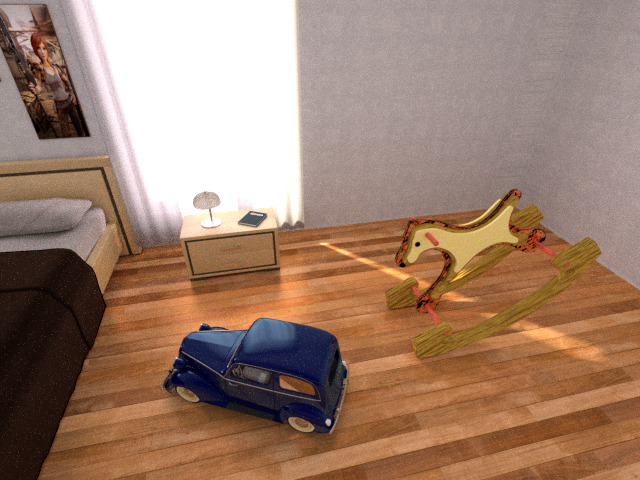} \\[5pt]
	\includegraphics[width=\linewidth]{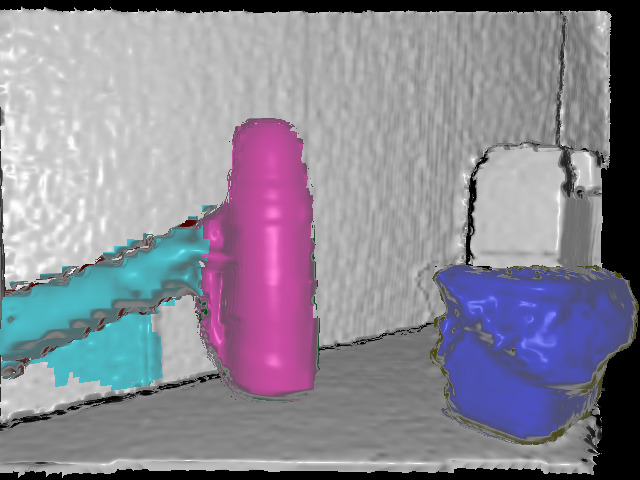} & \includegraphics[width=\linewidth]{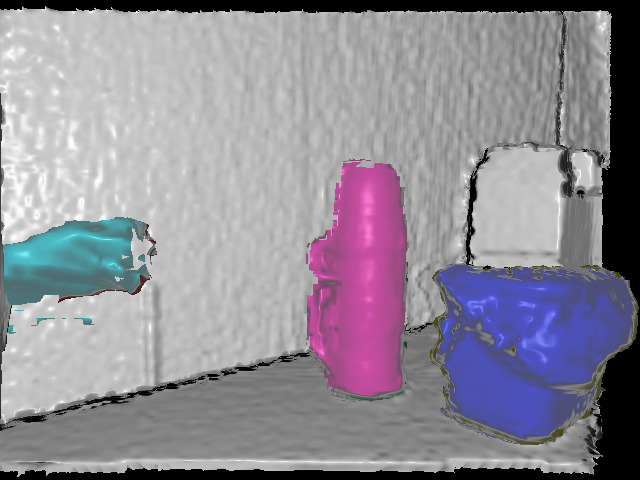} & \includegraphics[width=\linewidth]{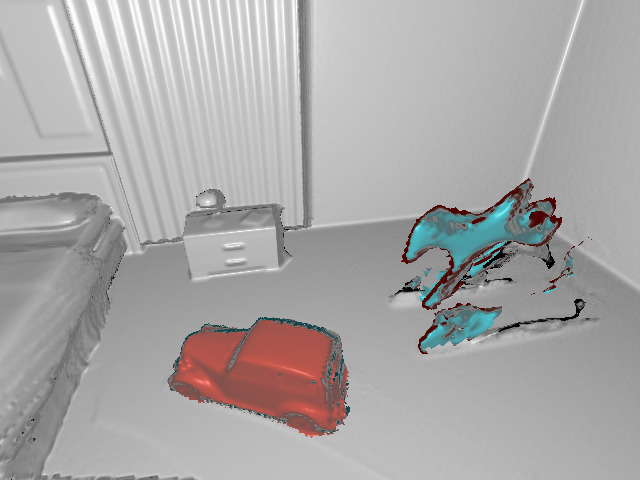} \\
	\includegraphics[width=\linewidth]{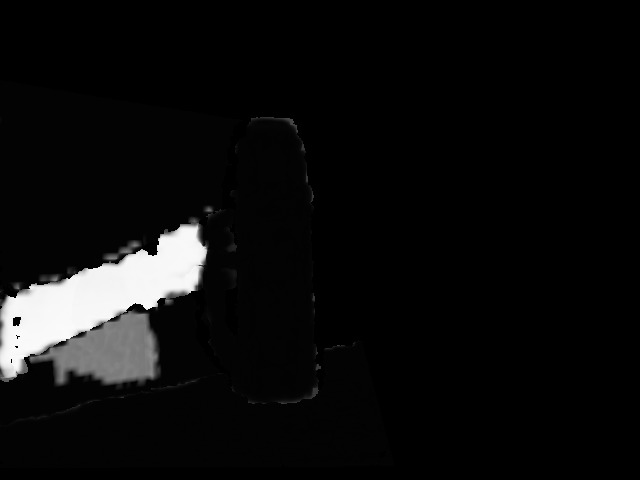} & \includegraphics[width=\linewidth]{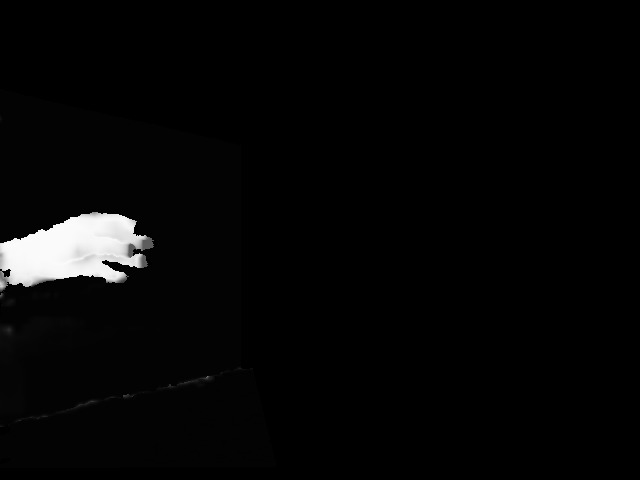} & \includegraphics[width=\linewidth]{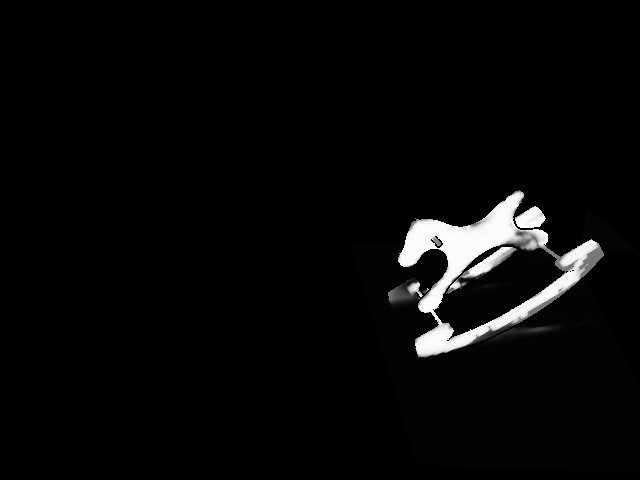} \\
	\includegraphics[width=\linewidth]{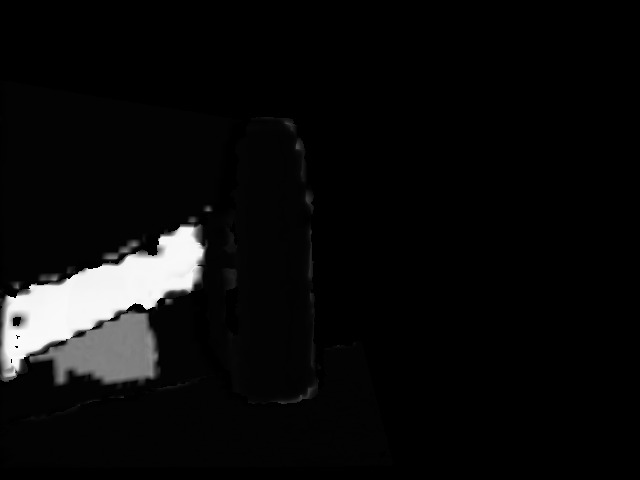} & \includegraphics[width=\linewidth]{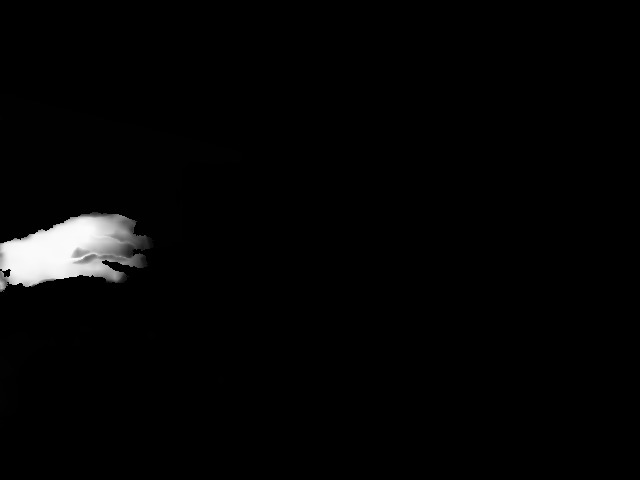} & \includegraphics[width=\linewidth]{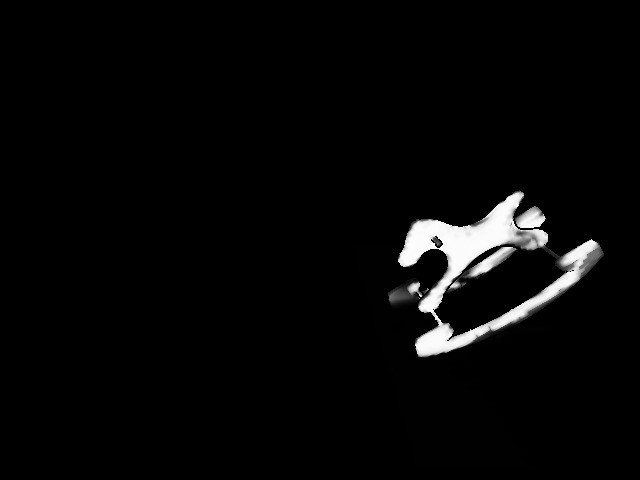} \\[5pt]
	
	\includegraphics[width=\linewidth]{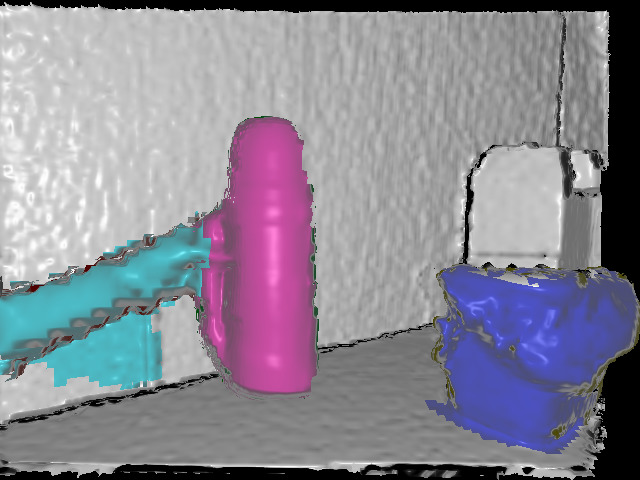} & \includegraphics[width=\linewidth]{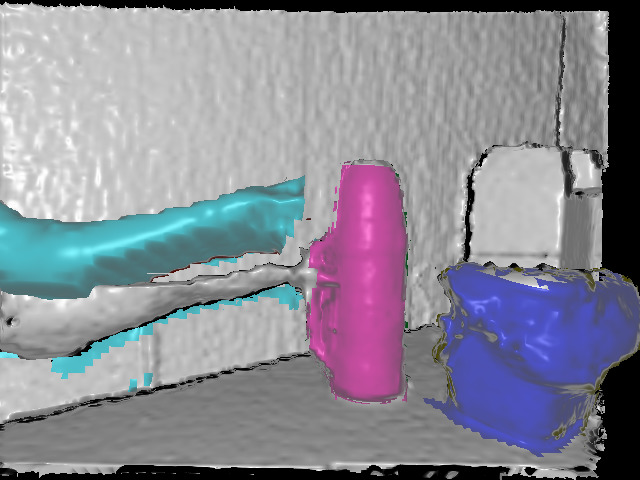} & \includegraphics[width=\linewidth]{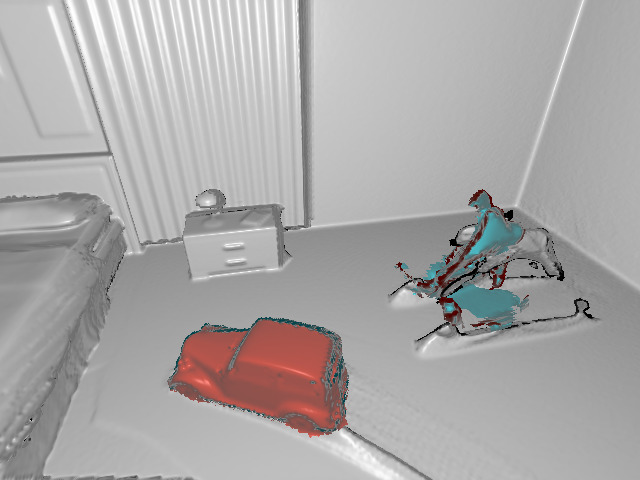} \\
	\includegraphics[width=\linewidth]{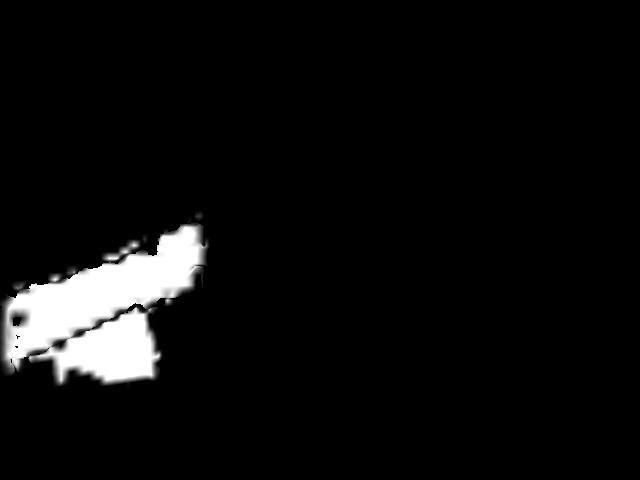} & \includegraphics[width=\linewidth]{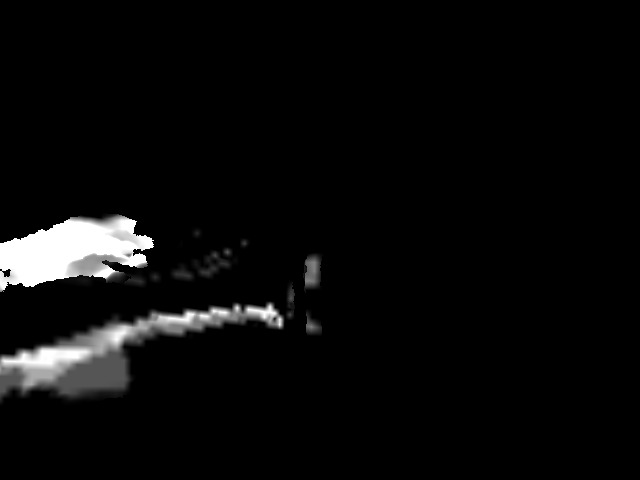} & \includegraphics[width=\linewidth]{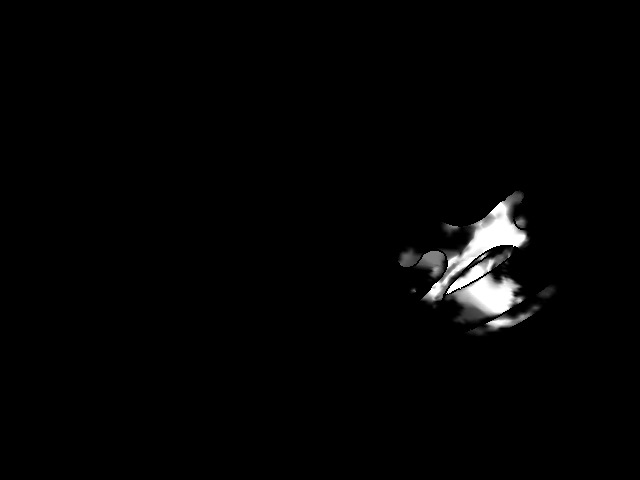} \\
	\end{tabular}
	\end{center}
	\caption{Tracking with association likelihoods. Probabilistic data association helps to overcome inaccuracies of the instance segmentation with geometric cues and makes the tracking more robust. From top to bottom: RGB images, our 3D reconstruction with reprojected object segmentation, association likelihood for the hand/horse object, our total pixel weights for tracking for the hand/horse object, 3D reconstruction with foreground probability instead of the association likelihood, total tracking weights with foreground probability instead of association likelihood.}
	\label{fig:tracking_assocweights}
\end{figure}

Most existing approaches to dynamic multi-object SLAM employ a variant of the iterative closest points (ICP~\cite{besl1992_icp}) algorithm for tracking the camera pose. 
This requires that a point cloud is extracted from the existing TSDF volume and associations are found between this point cloud and the depth image.
A typical approach with SDF map representations is to apply raycasting to determine the zero-crossings along the line-of-sight of the pixels.
The point clouds are aligned using non-linear least squares techniques.
In this approach, depth measurements are associated projectively with the zero-level surface.

We instead follow the approach in~\cite{bylow2013_sdftracking} and associate the depth measurements with the closest point on the surface.
This is achieved by minimizing the signed distance of the measured points to the surface which is directly given by the SDF function at the points.
The main advantage of this strategy is that pixels are associated with the correct part of the implicit surface using only one trilinear interpolation lookup per pixel in each iteration of the algorithm.
In ICP, the projective association is only performed once and requires several lookups per pixel until a surface is found.

For the M-step in Eq.~\eqref{eq:mstep}, we estimate the camera pose with regard to an SDF volume by minimizing
\begin{equation}
\label{eq:trackingerror}
	E( \vecsym{\xi} ) = \frac{1}{2} \sum_{\vec{u} \in \matsym{\Omega}} q( c_u ) \left| \psi\left( \mat{T}( \vecsym{\xi} ) \, \overline{\vec{p}}(\vec{u}) \right) \right|_\delta,
\end{equation}
where $\vec{p}(\vec{u}) := \pi^{-1}( \vec{u}, D(\vec{u}) )$ and $q( c_u )$ is the association likelihood of pixel $\vec{u}$ for the object/background.
We use the Huber norm with threshold $\delta$ to achieve robustness with regard to outliers.

We optimize Eq.~\eqref{eq:trackingerror} using the iteratively reweighted non-linear least squares (IRLS) algorithm.
Since the camera poses are in~$\SEthree$, we optimize Eq.~\eqref{eq:trackingerror} by reformulating it with a local parametrization using the Lie algebra~$\se(3)$. 
To this end, we apply local increments~$\vecsym{\delta \xi} \in \se(3)$ to the current solution for $\vecsym{\xi}$ in each iteration which we linearize at~$\vecsym{\delta \xi} = \vec{0}$.
Consequently, Eq.~\eqref{eq:trackingerror} becomes
\begin{equation}
	\label{eq:trackingerror_weighted}
	E( \vecsym{\delta\xi} ) = \frac{1}{2} \sum_{\vec{u} \in \matsym{\Omega}} q( c_u ) \, w_u \left( \psi\left( \mat{T}( \vecsym{\xi} ) \mat{T}( \vecsym{\delta \xi} ) \, \overline{\vec{p}}(\vec{u}) \right) \right)^2,
\end{equation}
with weights~$w_u$ which are adapted in each iteration to implement the Huber norm.
We additionally weigh the individual terms in the sum in \eqref{eq:trackingerror_weighted} by the map confidence $W(\pi^{-1}(\vec{u} ,D(\vec{u})))/ \max_{\vec{u}'\in\matsym{\Omega}} W(\pi^{-1}(\vec{u}' ,D(\vec{u}')))$, where $W(\vec{p})$ is the accumulated integration weight (see section \ref{sec:mapping}). 
It quantifies how certain we are about a surface estimate in the model.
This robustifies the tracking when large objects enter the frame from the image boundary. 
The optimization is performed using the Levenberg-Marquardt method.
This tracking optimization is first run on the background TSDF to estimate the updated camera pose before recomputing the association probabilities and running the same algorithm on each object TSDF for updating the individual object poses.

Fig.~\ref{fig:tracking_assocweights} illustrates the effectiveness of using the association likelihood for tracking.
We compare our approach with just using the foreground probabilities without geometric cues by replacing $q(c_u)$ with $p_\mathit{fg}(\vec{p}_{c_u} \mid c_u)$ in Eq.\ \eqref{eq:trackingerror_weighted}.
While the foreground probability also provides a segmentation cue, it is not sufficient for robust tracking due to the inaccurate instance segmentations by Mask R-CNN.

\subsection{Mapping}\label{sec:mapping}

Once the new camera poses~$\vecsym{\xi}_t$ have been estimated, we implement the M-step (Eq.~\eqref{eq:mstep}) by integrating the depth maps into the background and object volumes.
Following~\cite{curless96_volsdf}, we find the SDF as the maximum likelihood surface fit to the depth images using the recursive integration
\begin{equation}
\begin{split}
	\psi( v ) &\leftarrow \frac{W( v ) \psi( v ) + q( c_u ) \, d( v )}{W( v ) + q( c_u )},\\
	W( v ) &\leftarrow \min( W_{max}, W( v ) + q( c_u ) ),
\end{split}
\end{equation}
where~$d(v)$ is the measured depth difference of the voxel towards the integrated depth image.
For implementing the M-step in Eq.~\eqref{eq:mstep}, we incorporate the association likelihood $q(c_u)$ of the pixel $\vec{u}$ which passes through the voxel for computing the update weight.
The cap on $W(v)$ prevents the model from becoming overconfident in the SDF estimate and allows for faster adaptation in case of inaccurate or missing segmentations of dynamic objects.
Non-moving objects are initially integrated in the background map as well until the moving object map fits the measurements better.
We consider backtracing these insertions too costly.
One could reweigh the accumulated weight $W(v)$ with $w_u$ for faster map updates which increases drift though.

\section{Experiments}

We evaluate the performance of our method qualitatively and quantitatively on datasets containing dynamic scenes published with \cite{ruenz2017_cofusion} and the benchmark \cite{sturm12iros}.
We employ the Mask R-CNN implementation of \cite{matterport_maskrcnn_2017}.
In our experiments, the truncation distance is chosen to be $10$ times the voxel size for each TSDF volume and the parameter $\delta$ in \eqref{eq:trackingerror} is twice the voxel size. In \eqref{eq:data_association}, we set $\sigma = 0.02$, $\alpha = 0.8$, and $p_\mathcal{U}(\vec{p}_{c_t}) = 1.0$.
Mask R-CNN detections are only accepted if they are large enough (at least $40 \times 40$ pixels) and objects are classified as invisible (tracking and mapping unreliable) and deleted if their projected mask area within a region of 20 pixels from the image boundary is below this threshold.
To avoid cluttering the scene with large volumes containing static objects for which Mask R-CNN usually generates very inaccurate masks, we exclude a list of these object classes (\eg, tables, beds, refrigerators, etc.) from the detections used for instantiating new object volumes.
While one could implement a sliding window version for the background TSDF \cite{whelan2012_kintinuous}, we found that in our experiments a volume size of $5.12$m with the camera positioned at the center of one of the sides of the volume usually worked well. The only exception from this strategy is the scene \emph{Room4}, where we increased the volume size to $7.68$m and moved the initial camera pose further inside the volume to keep the scene within the volume boundaries.

\subsection{Quantitative Evaluation}

\textbf{Tracking of dynamic objects.} We perform quantitative evaluation of dynamic object tracking on the synthetic scenes provided by the authors of Co-Fusion \cite{ruenz2017_cofusion}.
Remarkably, although many objects present in the scenes are not contained in the COCO dataset~\cite{lin2014_coco}, Mask R-CNN manages to generate detections of most of the moving objects.
We compare our method to Kintinuous (KT,~\cite{whelan2012_kintinuous}), ElasticFusion (EF,~\cite{whelan2015_elasticfusion}), Co-Fusion (CF,~\cite{ruenz2017_cofusion}), and MaskFusion (MF,~\cite{ruenz2018_maskfusion}). 
KT and EF are static SLAM systems that treat dynamic objects as outliers.
CF uses geometric and motion segmentation for dynamic objects, while MF combines geometric segmentation with Mask R-CNN based instance segmentation.
For the publicly available implementation of MF we adjusted the minimum number of pixels required for instantiating a new object model to work well on the sequences.
We used the same threshold as in our approach, but MF still failed to instantiate an object instance for the rocking horse in the \emph{Room4} scene.

The results of our evaluation are shown in Table~\ref{tab:co-fusion-comp}. 
One can see that our method achieves competitive results.
Especially for the dynamic objects, our method outperforms the competing dynamic object-level SLAM approaches CF and MF.
The large camera tracking error wrt the static background (Static Bg) for MF in the \emph{ToyCar3} scene is caused by a very late detection of one of the moving cars, causing significant drift at the beginning of the trajectory.
This shows that the ICP tracking without a robust norm used in MF is sensitive to missing detections.
Our robust tracking using direct SDF alignment and the Huber norm, however, manages to keep the trajectory error low.

\begin{table}
\footnotesize
	\begin{center}
	\begin{tabular}{llccccc}
	\toprule
		&& KT & EF & CF & MF & \bf Ours \\
		\midrule
		\multirow{3}{*}{\rotatebox{90}{\emph{ToyCar3}}} & Static Bg & \textbf{0.10} & 0.59 & 0.61 & 20.60 & 0.95 \\
		& Car1 & - & - & 7.78 & 1.53 & \textbf{0.77} \\
		& Car2 & - & - & 1.44 & 0.58 & \textbf{0.18} \\
		\midrule
		\multirow{6}{*}{\rotatebox{90}{\emph{Room4}}} & Static Bg & \textbf{0.16} & 1.22 & 0.93 & 1.41 & 1.37 \\
		& Airship & - & - & 0.91/ & 13.62/ & \textbf{0.56}/ \\
		&&&& 1.01 & 2.29/ & 1.41/ \\
		&&&& & 3.46 & 0.75 \\
		& Car & - & - & \textbf{0.29} & 2.66 & 2.10 \\
		& Horse & - & - & 5.80 & - & \textbf{3.57} \\
		\bottomrule
	\end{tabular}
	\end{center}
	\caption{AT-RMSEs (in cm) of estimated trajectories for the synthetic sequences from Co-Fusion \cite{ruenz2017_cofusion}. The Airship trajectory is split into multiple parts due to separate geometric segments and detections with too little overlap for assignment. Our method achieves competitive results with a static SLAM system (EF) for the static background and outperforms other dynamic SLAM approaches (CF, MF) on the objects.}
	\label{tab:co-fusion-comp}
\end{table}

\begin{table}
\footnotesize
	\begin{center}
	\begin{tabular}{lcccccc}\toprule
		& VO-SF & SF & CF & MF & MID-F & \bf Ours \\
		\midrule
		f3s static & 2.9 & 1.3 & 1.1 & 2.1 & 1.0 & \textbf{0.9} \\
		f3s xyz & 11.1 & 4.0 & \textbf{2.7} & 3.1 & 6.2 & 3.7 \\
		f3s halfsphere & 18.0 & 4.0 & 3.6 & 5.2 & \textbf{3.1} & 3.2 \\
		f3w static & 32.7 & \textbf{1.4} & 55.1 & 3.5 & 2.3 & \textbf{1.4} \\
		f3w xyz & 87.4 & 12.7 & 69.6 & 10.4 & 6.8 & \textbf{6.6} \\
		f3w halfsphere & 73.9 & 39.1 & 80.3 & 10.6 & \textbf{3.8} & 5.1 \\
		\bottomrule
	\end{tabular}

	(a) Absolute trajectory (AT) RMSE (in cm)\\
	\ \\

	\begin{tabular}{lccccc}
		\toprule
		& VO-SF & CF & SF & MF &  \bf Ours \\
		\midrule
		f3s static & 2.4 & 1.1 & 1.1 & 1.7 & \textbf{0.9} \\
		f3s xyz & 5.7 & 2.7 & 2.8 & 4.6 & \textbf{2.6} \\
		f3s halfsphere & 7.5 & \textbf{3.0} & \textbf{3.0} & 4.1 & \textbf{3.0} \\
		f3w static & 10.1 & 22.4 & 1.3 & 3.9 & \textbf{1.2} \\
		f3w xyz & 27.7 & 32.9 & 12.1 & 9.7 & \textbf{6.0} \\
		f3w halfsphere & 33.5 & 40.0 & 20.7 & 9.3 & \textbf{5.1} \\
		\bottomrule
	\end{tabular}

	(b) Relative pose (RP) RMSE (cm/s)\\
	\end{center}
	
	\caption{Comparison of robust camera tracking wrt. the static background in dynamic scenes for different methods. Our approach provides state-of-the-art results and outperforms previous methods in the majority of sequences.}
	\label{tab:tum-comp}
\end{table}

\begin{table}
	\footnotesize
	\begin{center}
	\begin{tabular}{llccc}
		\toprule
		&& w/o assoc. & w/o map conf.  & Ours \\
		\midrule
		\multirow{6}{*}{\rotatebox{90}{\emph{Room4}}} & Static Bg & 1.42 & \bf 1.37 & \bf 1.37 \\
		& Airship & \bf 0.49/ & 0.73/ & 0.56/ \\
		&&          \bf 1.13/ & 1.47/ & 1.41/ \\
		&&          1.24  & \bf 0.75 & \bf 0.75 \\
		& Car & \bf 2.01 & 2.11 & 2.10 \\
		& Horse & 9.12 & 8.38 & \bf 3.57 \\
		\bottomrule
	\end{tabular}
	\end{center}
	\caption{Ablation study on the synthetic scene \emph{Room4}. We compare AT-RMSE for our approach to not using association likelihoods, and to not using map confidence weights for tracking.}
	\label{tab:ablation}
\end{table}

\begin{figure*}
\footnotesize
	\begin{center}
	\setlength{\tabcolsep}{1pt}
	\renewcommand{\arraystretch}{0.6}
	\begin{tabular}{p{.15\linewidth}p{.15\linewidth}p{.15\linewidth}p{.15\linewidth}p{.15\linewidth}}
		\includegraphics[width=\linewidth]{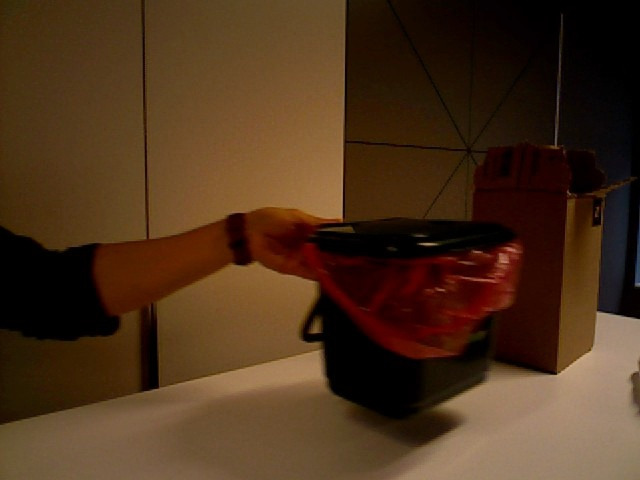} & \includegraphics[width=\linewidth]{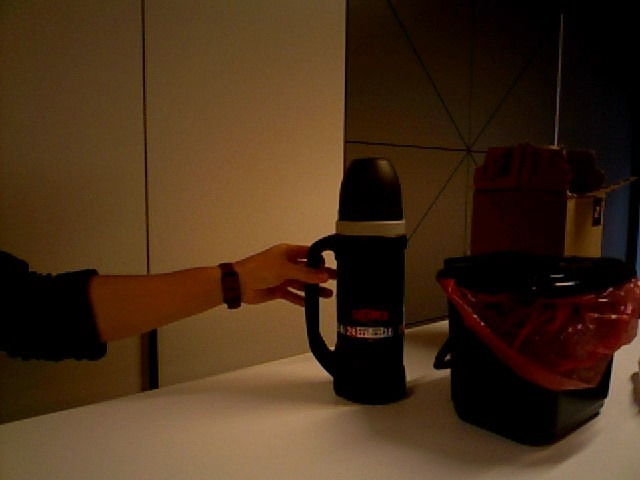} & \includegraphics[width=\linewidth]{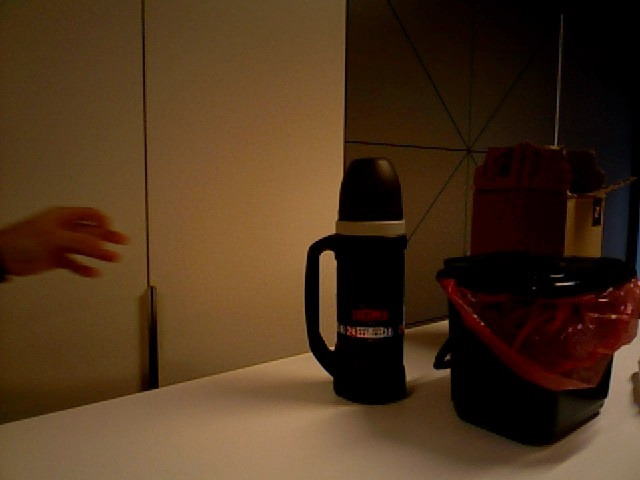} & \includegraphics[width=\linewidth]{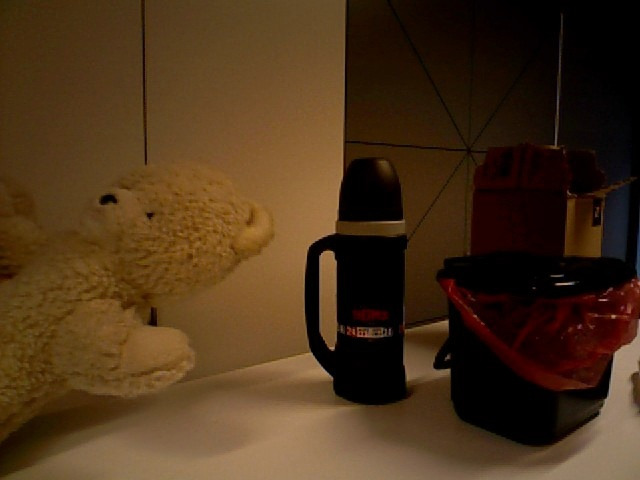} & \includegraphics[width=\linewidth]{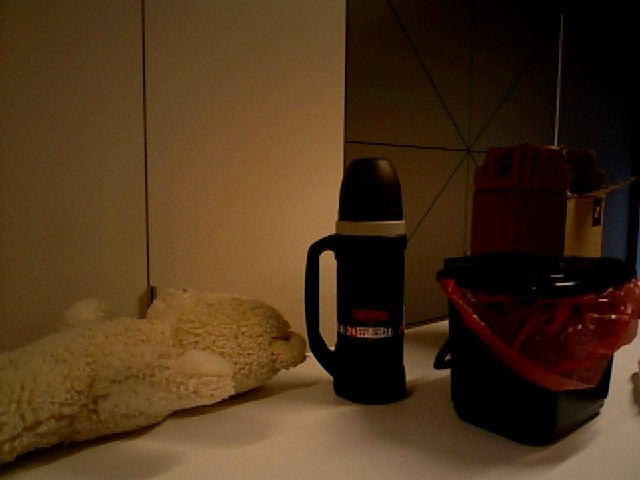} \\
		\includegraphics[width=\linewidth]{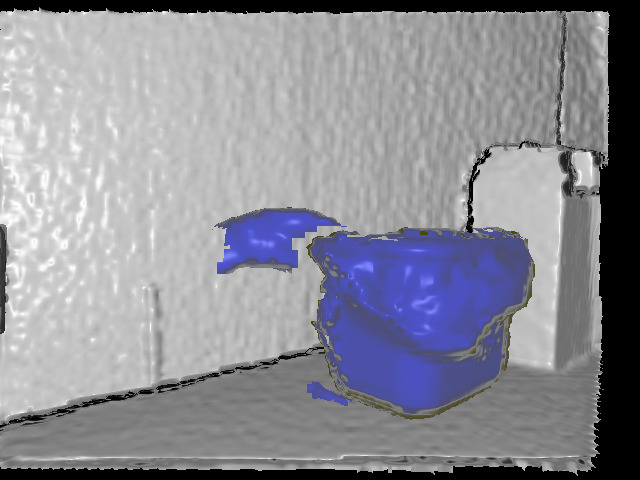} & \includegraphics[width=\linewidth]{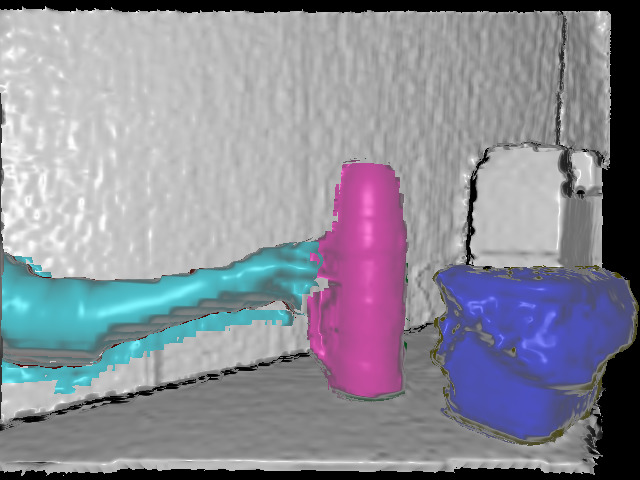} & \includegraphics[width=\linewidth]{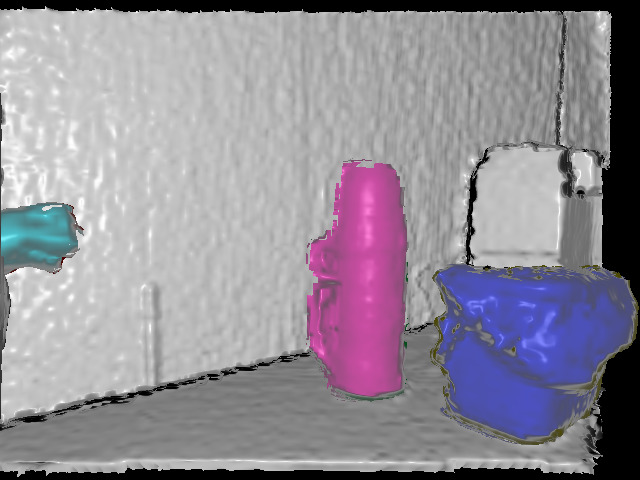} & \includegraphics[width=\linewidth]{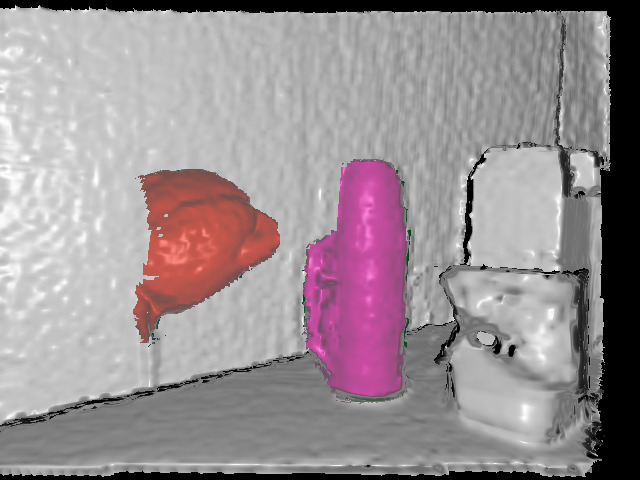} & \includegraphics[width=\linewidth]{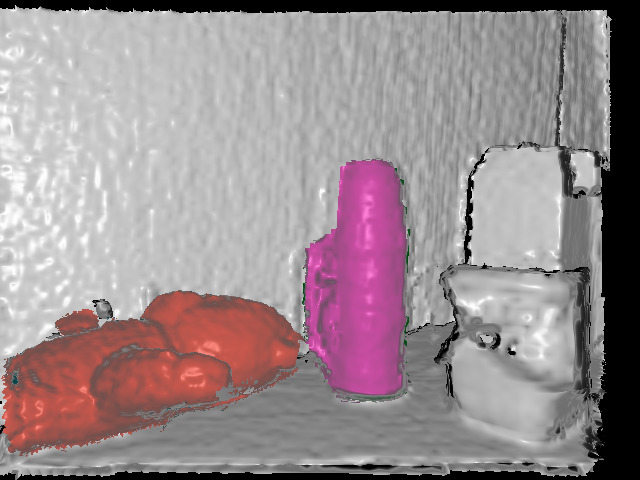} \\
		\includegraphics[width=\linewidth]{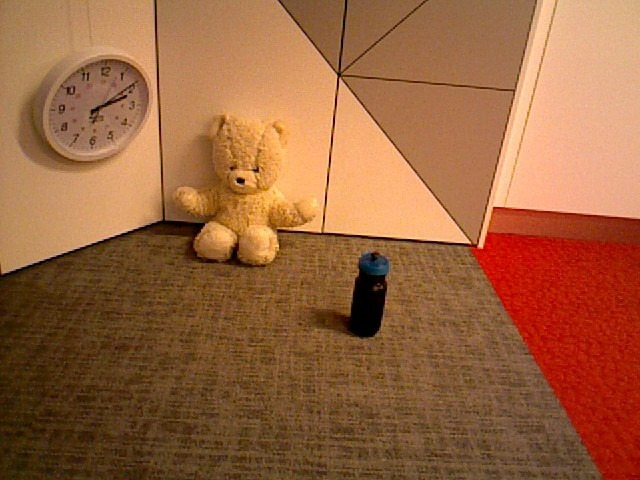} & \includegraphics[width=\linewidth]{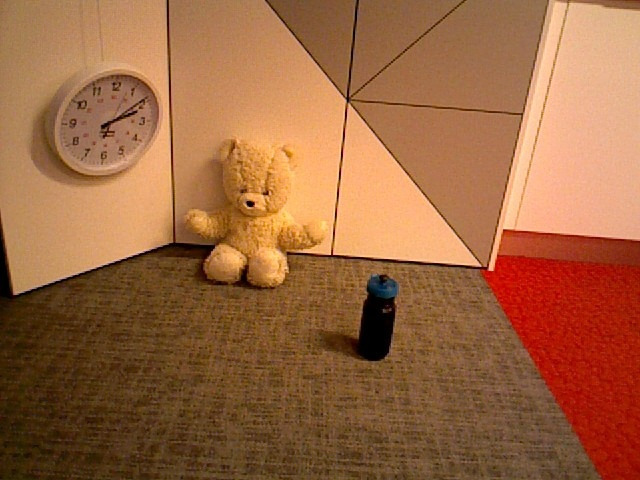} & \includegraphics[width=\linewidth]{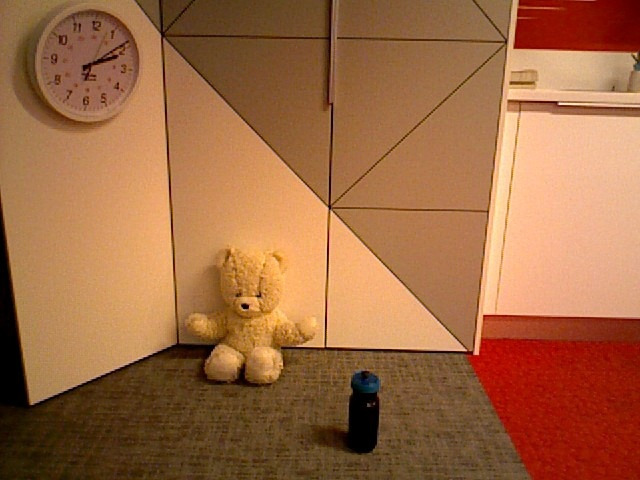} & \includegraphics[width=\linewidth]{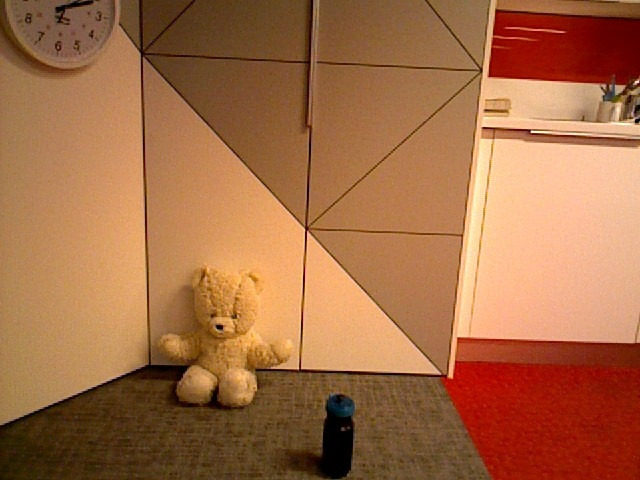} & \includegraphics[width=\linewidth]{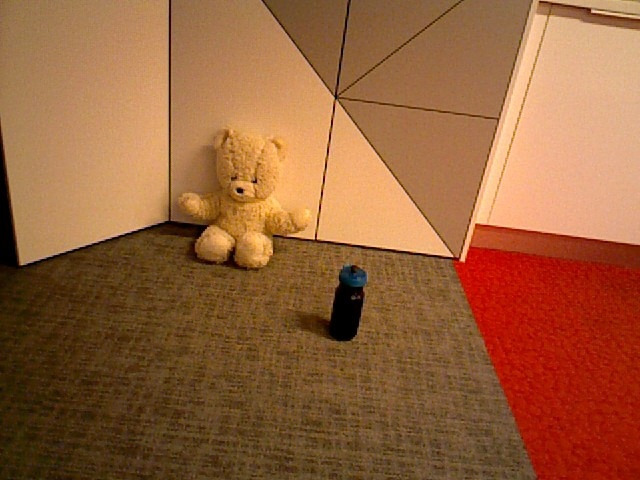} \\
		\includegraphics[width=\linewidth]{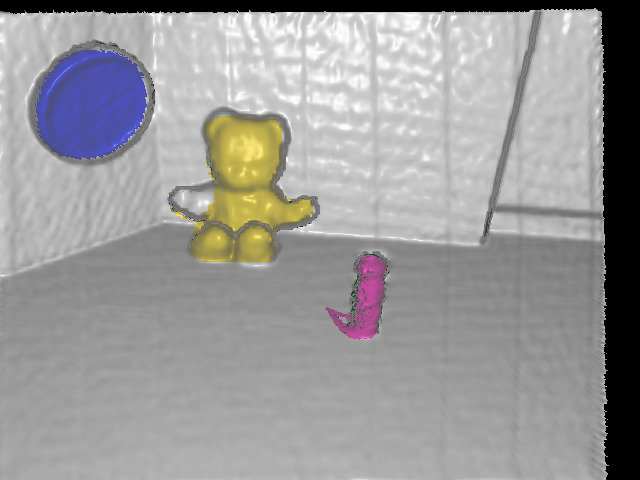} & \includegraphics[width=\linewidth]{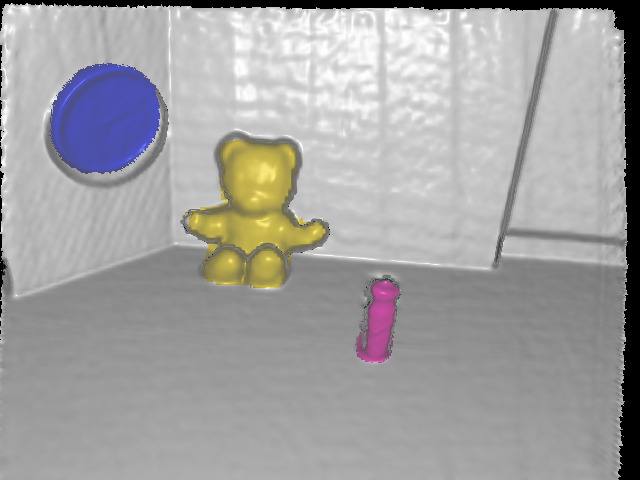} & \includegraphics[width=\linewidth]{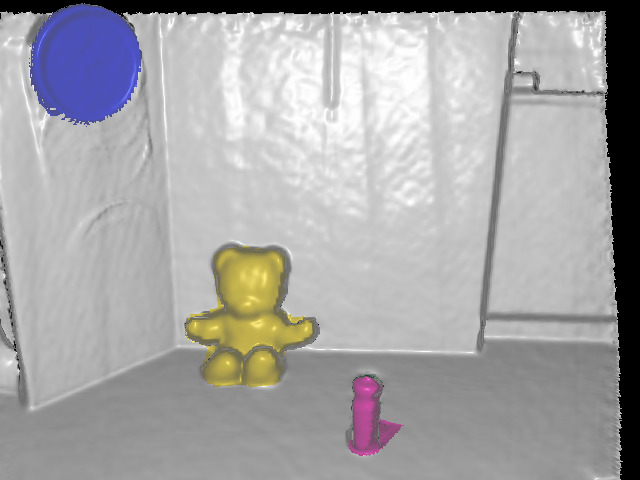} & \includegraphics[width=\linewidth]{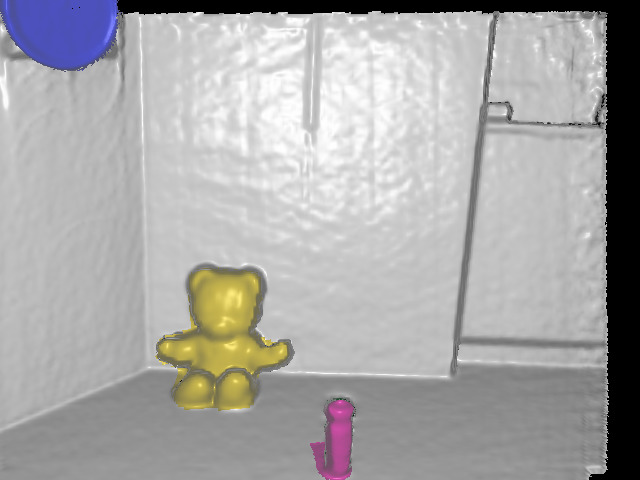} & \includegraphics[width=\linewidth]{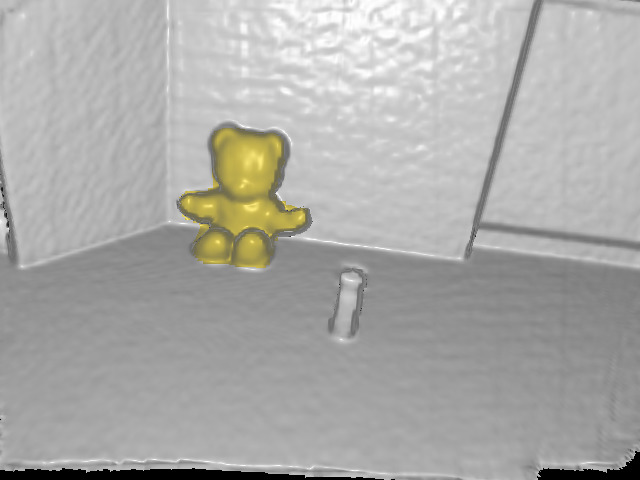} \\
		\includegraphics[width=\linewidth]{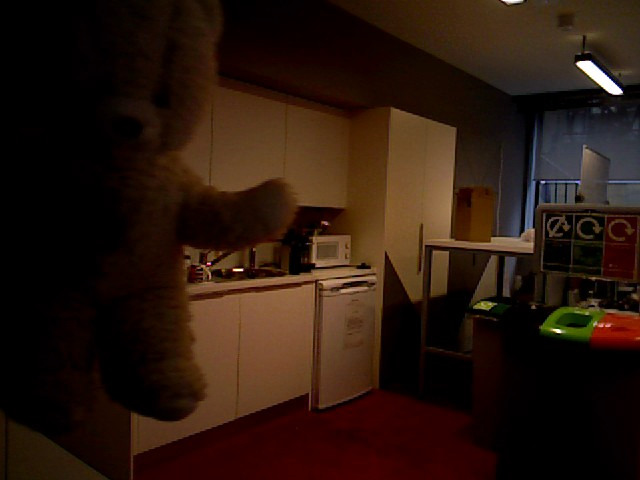} & \includegraphics[width=\linewidth]{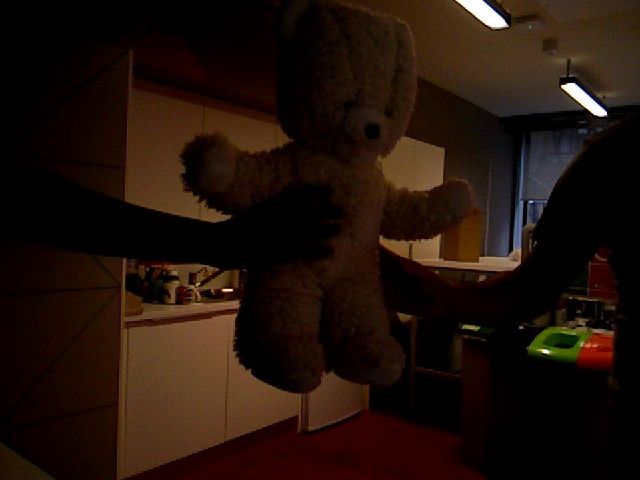} & \includegraphics[width=\linewidth]{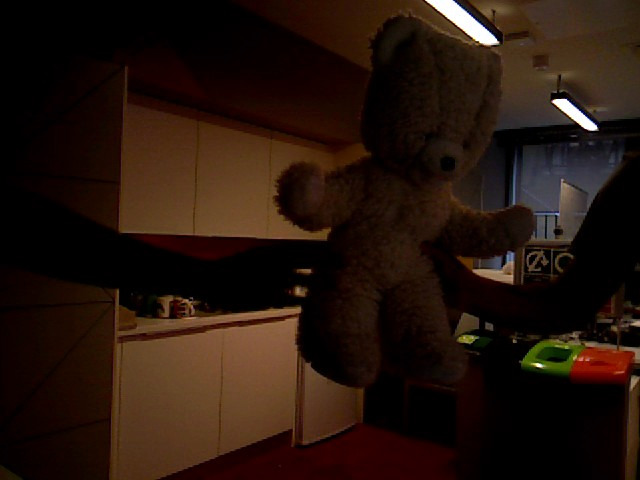} & \includegraphics[width=\linewidth]{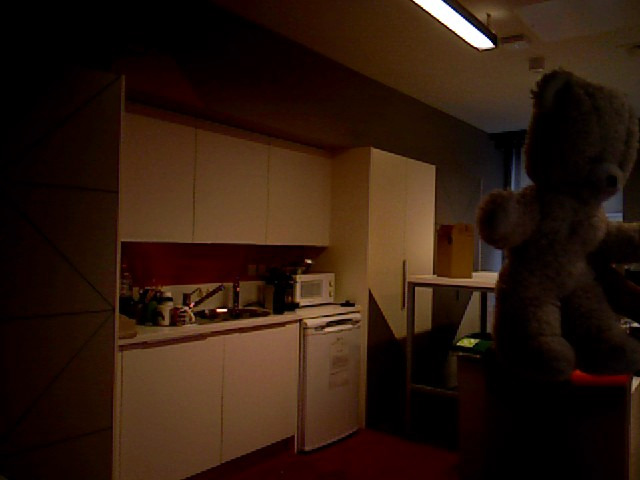} & \includegraphics[width=\linewidth]{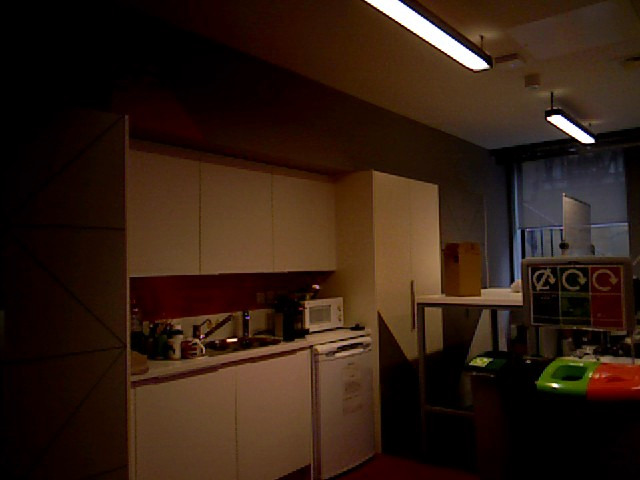} \\
		\includegraphics[width=\linewidth]{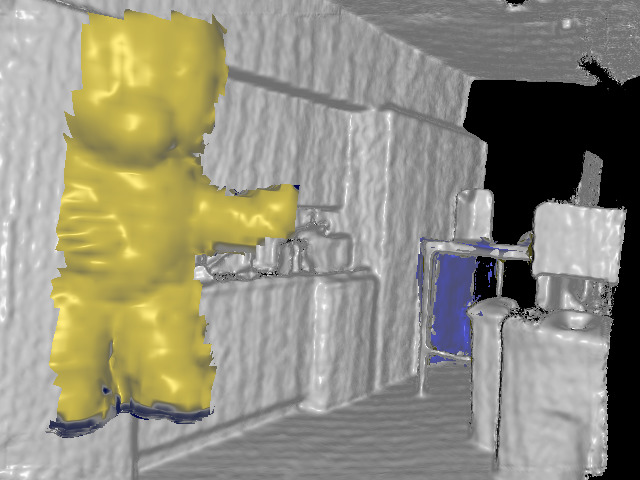} & \includegraphics[width=\linewidth]{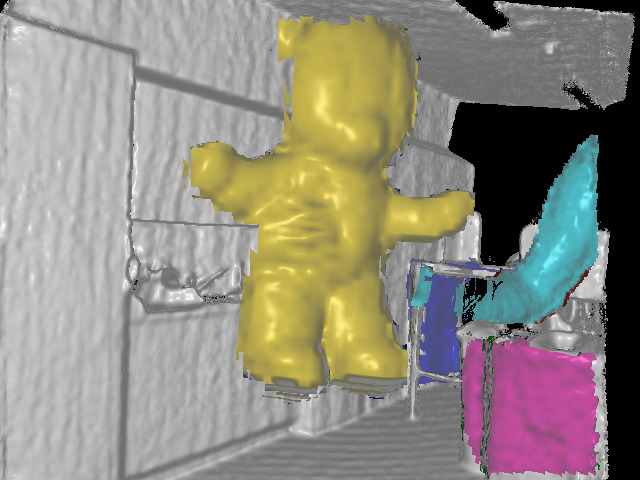} & \includegraphics[width=\linewidth]{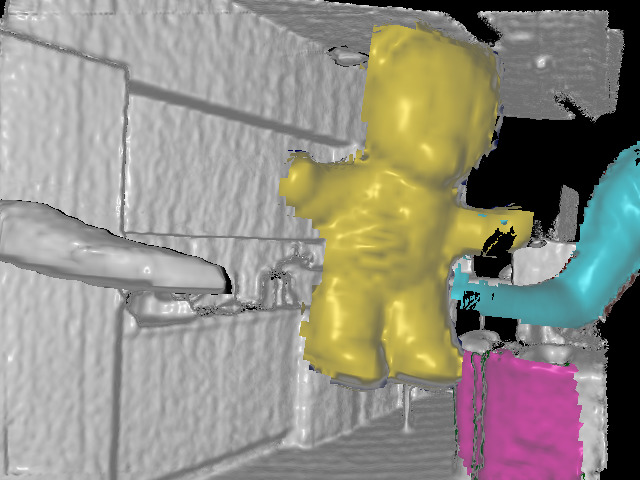} & \includegraphics[width=\linewidth]{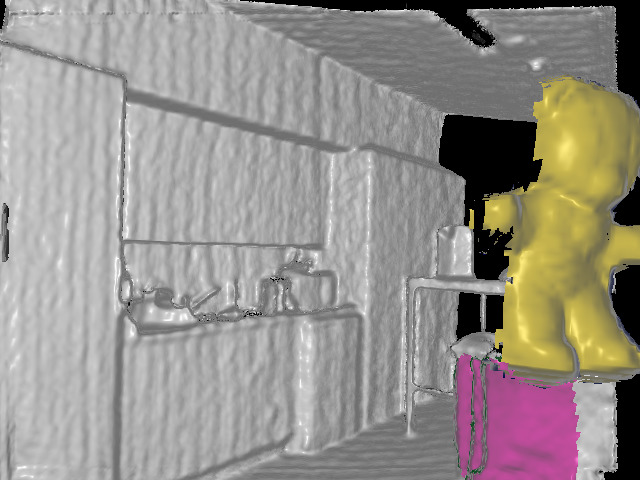} & \includegraphics[width=\linewidth]{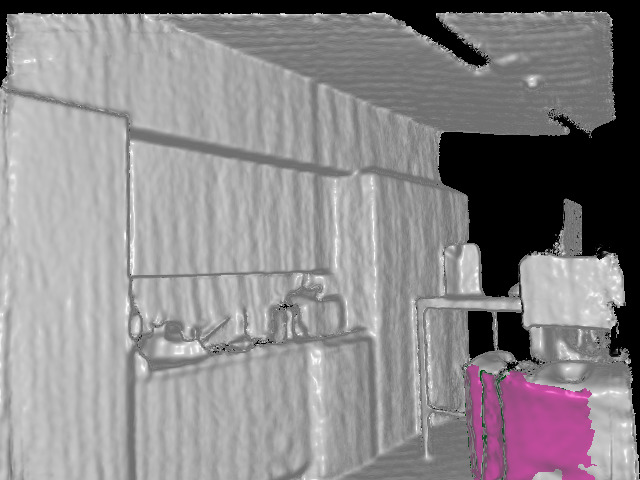} \\
	\end{tabular}
	\end{center}
	\caption{Qualitative evaluation on the real-world datasets published with Co-Fusion \cite{ruenz2017_cofusion}. We demonstrate that we can handle fast movement (the second and the third image of the first dataset only 25 frames apart), as well as objects with relative weak geometric cues, such as the clock in the second dataset. Note that the left arm handing over the teddy is not detected in the last dataset. While it initially is integrated into the background it is quickly overriden by actual background depth soon after it moved out of view.}
	\label{fig:co-fusion-real}
\end{figure*}

\begin{figure}
\footnotesize
	\begin{center}
	\setlength{\tabcolsep}{1pt}
	\renewcommand{\arraystretch}{0.6}
	\begin{tabular}{P{.3\linewidth}P{.3\linewidth}P{.3\linewidth}}
		\includegraphics[width=\linewidth]{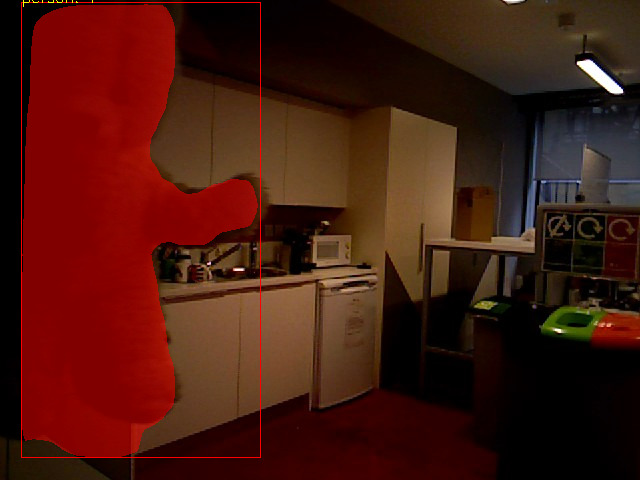} & 		\includegraphics[width=\linewidth]{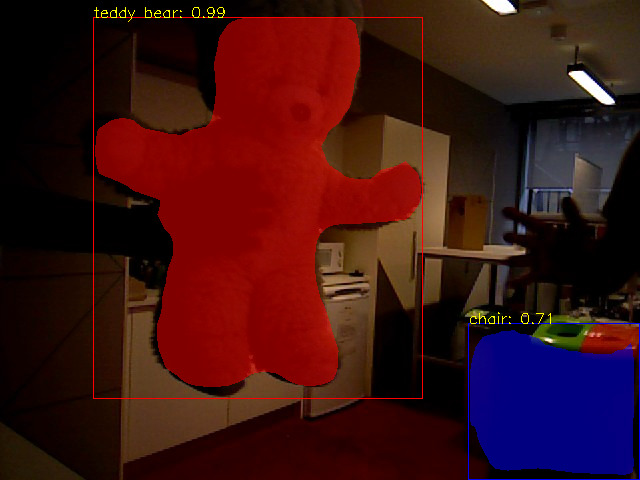} & 		\includegraphics[width=\linewidth]{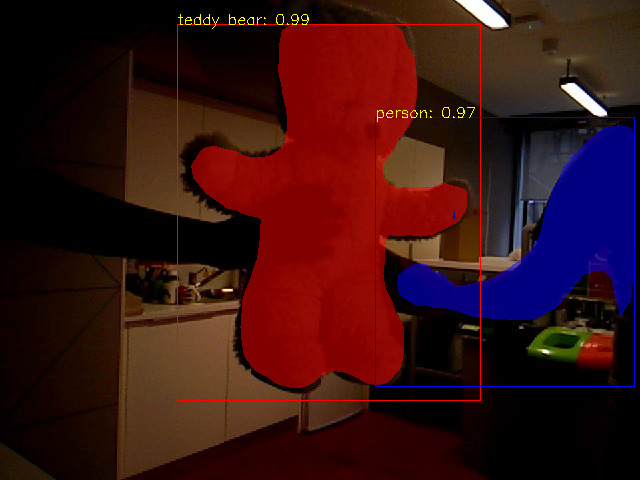} \\
		\includegraphics[width=\linewidth]{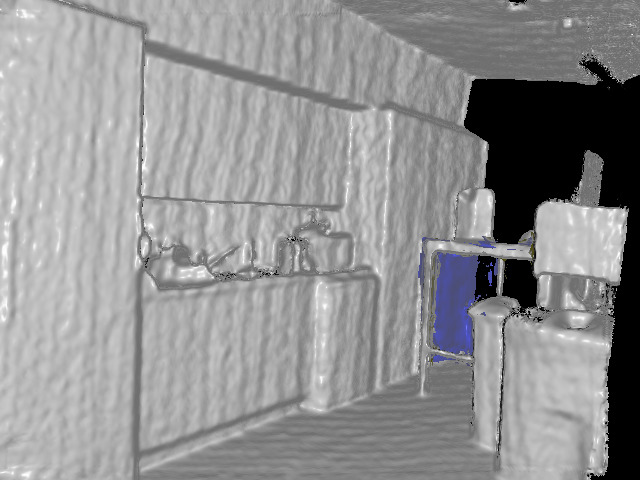} & 		\includegraphics[width=\linewidth]{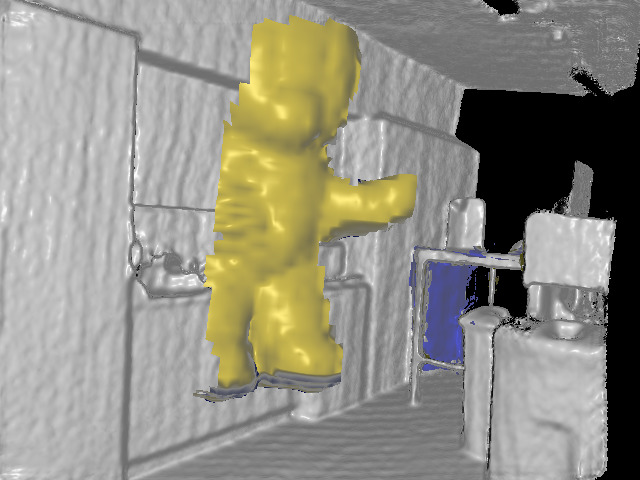} & 		\includegraphics[width=\linewidth]{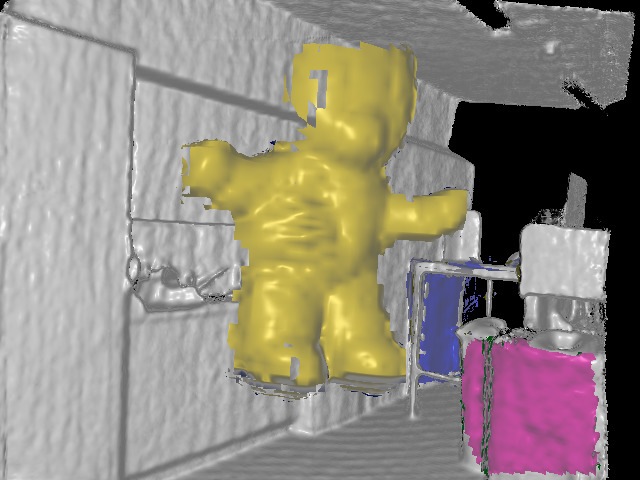} \\
		 & 		\includegraphics[width=\linewidth]{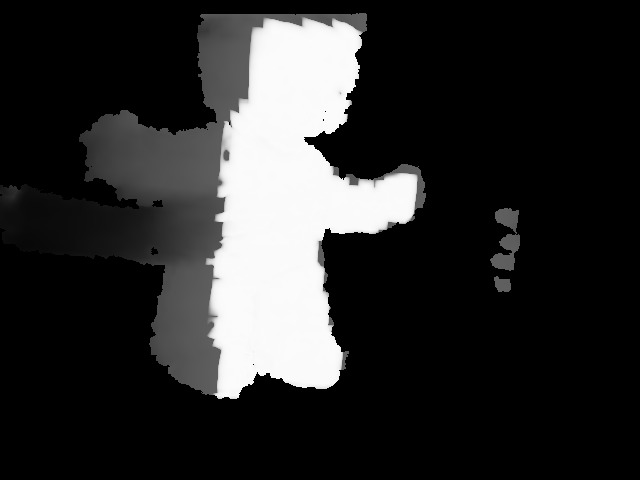} & 		\includegraphics[width=\linewidth]{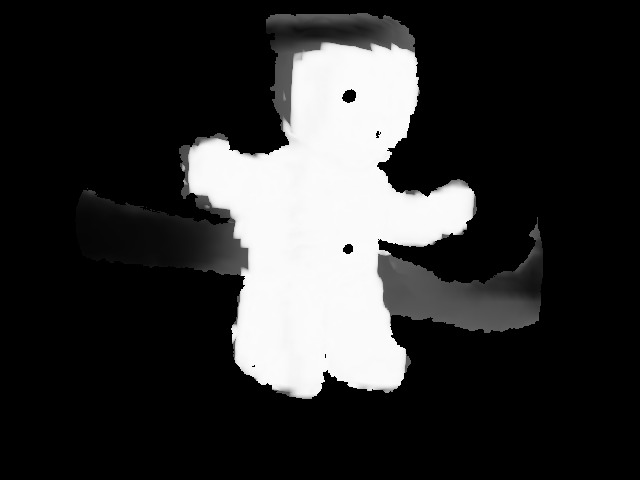} \\
		 		\includegraphics[width=\linewidth]{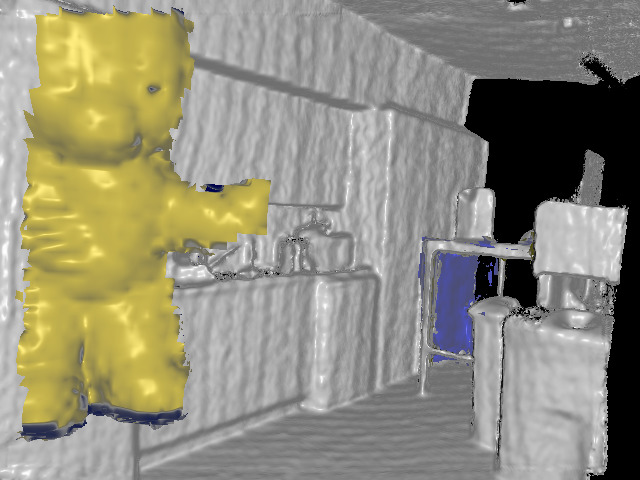} & 		\includegraphics[width=\linewidth]{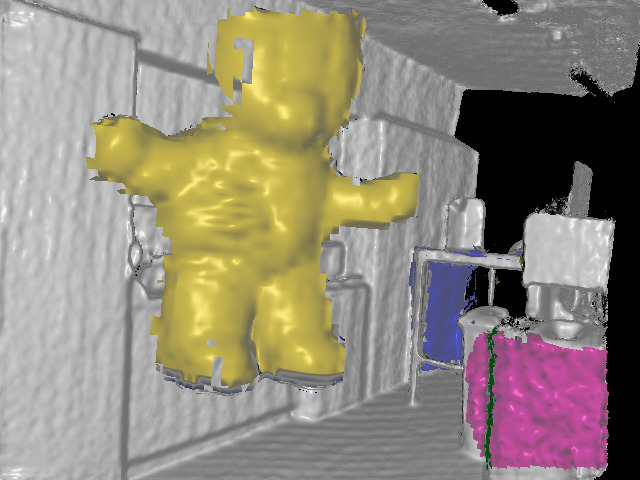} & 		\includegraphics[width=\linewidth]{images/teddy-handover/0301_output} \\
		\includegraphics[width=\linewidth]{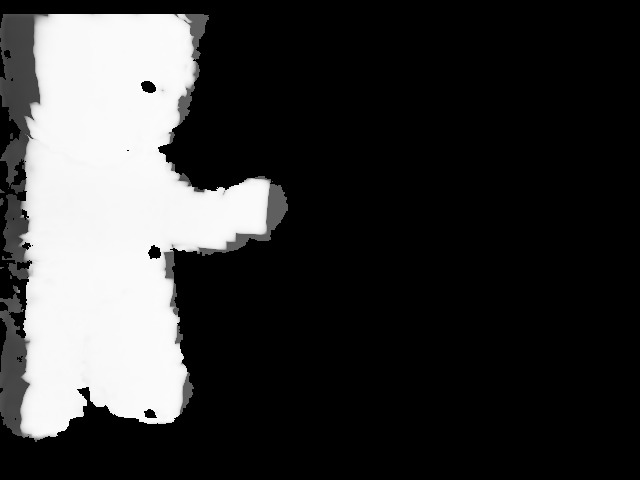} & 		\includegraphics[width=\linewidth]{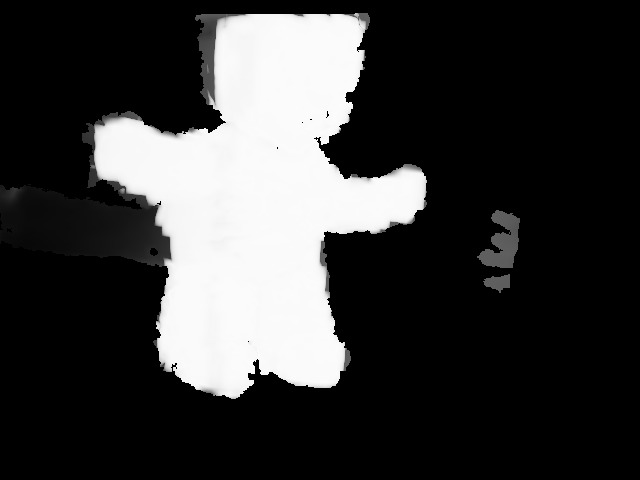} & 		\includegraphics[width=\linewidth]{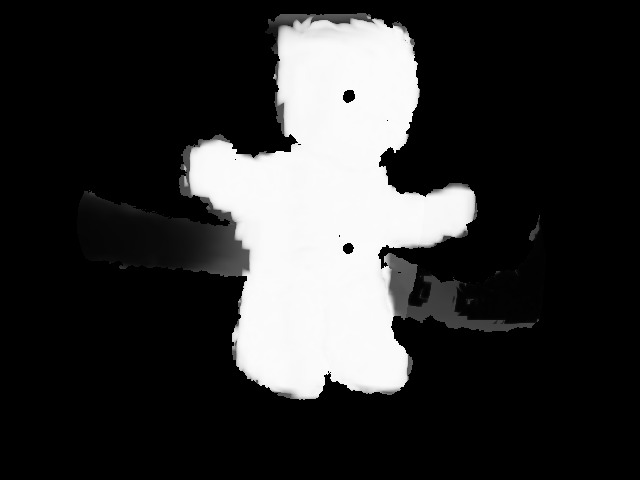} \\
		Frame 240 & Frame 270 & Frame 300
	\end{tabular}
	\end{center}
	\caption{Incremental mask integration. From top to bottom: Masked RGB Frame, model output and association likelihood for teddy before mask integration, model output and association likelihoods after mask integration. One can see that the association likelihoods provide a soft geometric segmentation for the moving geometry inside the volume of the teddy object. It gets stronger for the pixels that actually belong to the object once Mask R-CNN confirms those pixels to belong to that object. Note that the teddy bear is first detected in frame 240 and thus does not have association likelihoods in this frame yet.}
	\label{fig:mask_int_incr}
\end{figure}

\textbf{Robust camera tracking.} Similar to experiments performed in MaskFusion \cite{ruenz2018_maskfusion} and MID-Fusion \cite{xu2019_midfusion}, we can use Mask R-CNN detections with certain labels (\eg, \emph{person}) to exclude these labels from the reconstruction and tracking. In our approach, the association likelihoods already prevent parts of depthmaps projecting into foreground parts of object volumes from being integrated into the background volume used for camera tracking.
We thus maintain object volumes for detected people but do not render them during raycasting for visualization. 
The association likelihood then tends to associate even non-rigidly moving people to the object volumes rather than the background, enabling us to robustly track the camera wrt. background.

We compare our method to five state-of-the-art dynamic SLAM approaches in Table \ref{tab:tum-comp}. Two of these, joint visual odometry and scene flow (VO-SF,~\cite{jaimez2017_vosf}), and StaticFusion (SF,~\cite{scona2018_staticfusion}) were designed for reconstructing the static background while ignoring dynamic parts.
The remaining ones, CF~\cite{ruenz2017_cofusion}, MF~\cite{ruenz2018_maskfusion}, MID-Fusion (MID-F,~\cite{xu2019_midfusion}) were designed for multi-object reconstruction. The latter two of these methods, like our approach, use Mask R-CNN \cite{he2017_maskrcnn} detections for instantiating objects.
One can see that our method achieves competitive results in most cases, especially compared to MF \cite{ruenz2018_maskfusion} and MID-F \cite{xu2019_midfusion}.
Like all these methods, our method might fail if large undetected objects cover the major part of the image.
Our results demonstrate that the combination of robust tracking and our data association strategy improves robustness on these sequences.
The table rows are ordered approximately by scene difficulty, so the latter rows exhibit large dynamic parts with heavy occlusions. \emph{f3s} abbreviates \emph{freiburg3\_sitting} while \emph{f3w} stands for \emph{freiburg3\_walking}.
MID-F did not report RP-RMSE and thus is not shown in Table \ref{tab:tum-comp} (b).

We further compare to MF \cite{ruenz2018_maskfusion} on the scene \emph{f3\_long\_office\_household} of the benchmark \cite{sturm12iros}.
By exporting the relative trajectory of the teddy bear and the camera, we can compare the object trajectory to the ground truth camera trajectory as was done in \cite{ruenz2018_maskfusion}.
While we achieve slightly worse results on the teddy bear trajectory (3.5cm, while MF achieved 2.2cm), our camera trajectory is more accurate (5.0cm compared to 8.9cm for MF).
Note that while MF improved their camera trajectory wrt.\ the background to 7.2cm AT-RMSE when not tracking the teddy bear, we do not expect a notable change for this case in our approach since the teddy is implicitly reconstructed with partial association likelihood in the background and would be disassociated and removed from it if it started moving.

In Table \ref{tab:ablation}, we do an ablation study to evaluate the contributions of different parts of our method. Since most objects only observe minor changes in their local topology (the Airship moving freely in the air, the car driving on the ground), and there are no large objects moving into view from the edge of the image, the effects of not using association likelihoods or map confidence weights for tracking are numerically negligible for most objects. However, the rocking horse is subject to topology changes in its surrounding since wall and floor intersect the volume at different angles. We observe a significant improvement for this object in Tab.~\ref{tab:ablation}.

\textbf{Computational performance.} While our implementation is not yet tuned for runtime performance (\eg, parallel processing of objects), the average runtime per frame on the CF-datasets \cite{ruenz2017_cofusion} ranges from 106ms to 257ms on an Nvidia GTX 1080 Ti GPU with 11GB of memory and an Intel Xeon Silver 4112 CPU with 4 cores and 2.6 GHz.
A more detailed analysis separating the runtime on detection frames as well as an ablation study on how varying detection frequencies affect trajectory coverage and accuracy can be found in the supplemental material.

\subsection{Qualitative Evaluation}

Figure \ref{fig:co-fusion-real} shows a qualitative evaluation on the real-world datasets published with Co-Fusion \cite{ruenz2017_cofusion}. One can see, that we manage to reconstruct dynamic and static objects in these scenes if they are detected by Mask R-CNN. Note that some of the objects, like the trashcan in the first sequence are not contained in the set of classes that Mask R-CNN is trained on. Thus, the trashcan is not detected for a large number of frames and deleted because of a low existence probability $p_\mathit{ex}$. The bottle in the clock sequence is deleted after it is classified as ``not visible'' because it moves out of view and the number of pixels in view is too low.

We show how the incremental integration of foreground probabilities into object volumes improves the object masks in Figure \ref{fig:mask_int_incr}. 
Finally, for a qualitative evaluation of the effect of the association likelihood, we refer to Figure \ref{fig:teaser}, where moving objects leave a visible trace because their depth values are integrated into the background, and Figure \ref{fig:tracking_assocweights}, which shows that they help to improve the tracking quality by including geometric cues if Mask R-CNN segmentations do not fit the actual object shape.

\section{Conclusions}
In this paper we propose a novel probabilistic formulation for dynamic object-level SLAM with RGB-D cameras.
We infer the latent data association of pixels with the objects in the map concurrently with the maximum likelihood estimates of camera poses and maps.
The maps are represented as volumetric signed distance functions.
For tracking, our probabilistic formulation facilitates direct alignment of depth images with the SDF representation.
Our results demonstrate that proper probabilistic treatment of data associations is a key ingredient to robust tracking and mapping in dynamic scenes.
To the best of our knowledge, our approach is the first that considers EM for dynamic object-level SLAM with RGB-D cameras.

Note that our approach treats the detected objects models always as dynamic.
While our experiments have shown that their poses are stable in most settings for static objects, in future work an additional classification into static and dynamic objects might be developed to prevent drifting of static objects and to refine the camera pose by tracking it relative to the static object volumes.
This might prove beneficial since the object volumes usually exhibit a higher relative resolution.
In future work we further plan to integrate information from the RGB image for tracking to further increase the accuracy and robustness of the method in planar surfaces.
Furthermore, more efficient data structures and global graph optimization are interesting directions to further scale our approach.
Finally, we plan to investigate how our approach could be used on mobile manipulation platforms for the interactive perception of objects.

\vspace{1em}
\noindent {\large \textbf{Acknowledgements.}} We acknowledge support from the BMBF through the Tuebingen AI Center 
(FKZ: 01IS18039B) and Cyber Valley.
The authors thank the International Max Planck Research School for Intelligent Systems (IMPRS-IS)
for supporting Michael Strecke.

{\small
\bibliographystyle{ieee_fullname}
\bibliography{strecke19emfusion}
}

\end{document}